\documentclass{elsart}

\journal{International Journal of Approximate Reasoning}

\usepackage{amsmath,amsfonts,amsthm,latexsym,stmaryrd}
\usepackage{amssymb}

\usepackage{amscd, amsmath, amssymb, epic, eepic,  delarray,stmaryrd,mathbbol,amsbsy,amsfonts,bussproofs}

\usepackage{appendix}
\usepackage[utf8]{inputenc}
\usepackage{graphicx,epsfig,epsf,epsfig}
\usepackage{pstricks,pst-node,rotating}
\usepackage{color}
\usepackage{tikz}
\usepackage{tikz-cd}
\usepackage{pgf}
\usepackage[ps,curve,2cell]{xypic}
\usetikzlibrary{arrows,automata}
\tikzstyle{my loopup}=[->, to path={
 .. controls +(50:1) and +(100:1) .. (\tikztotarget) \tikztonodes}]

\tikzstyle{my loopdown}=[->, to path={
 .. controls +(-110:1) and +(-30:1) .. (\tikztotarget) \tikztonodes}]

\tikzstyle{my arcdown}=[->, to path={
 .. controls +(-100:1) and +(-120:1) .. (\tikztotarget) \tikztonodes}]

\tikzstyle{my arcup}=[->, to path={
 .. controls +(100:1) and +(120:1) .. (\tikztotarget) \tikztonodes}]

\usetikzlibrary{matrix}

\newcommand{\registeruser}[2]{%
  \expandafter\newcommand\csname #1\endcsname[1]{\textcolor{#2}{##1}}
  \newenvironment{#1-env}{\color{#2}}{}
}
\usepackage[normalem]{ulem}
\registeruser{Isa}{black}
\registeruser{Marc}{black}
\registeruser{Salim}{black}
\registeruser{Ramon}{black}

\newcommand{\contrac}{\ensuremath{ - \!\!\!\!  ^\cdot\; }}

\newcommand\Omit[1]{}

\begin{document}


\newcommand{\Dcal}{\mathcal{D}}

\newcommand{\Cat}{\mathrm{Cat}}
\newcommand{\Set}{\mathrm{Set}}
\newcommand{\Hey}{\mathrm{Hey}}
\newcommand{\Sub}{\mathrm{Sub}}
\newcommand{\Hom}{\mathrm{Hom}}
\newcommand{\nat}{\mathrm{Nat}}
\newcommand{\Sieve}{\mathrm{Sieve}}

\newcommand{\dom}{\mathrm{dom}}
\newcommand{\codom}{\mathrm{cod}}
\newcommand{\im}{\mathrm{Im}}

\newlength\height
\newlength\heightb
\newlength\depth
\newlength\depthb
\newcommand\restr[2]%
  {{%
  \settoheight\height{$\scriptstyle{#2}$}%
  \settodepth\depth{$\scriptstyle{#2}$}%
  \settodepth\depthb{$#1$}%
  \addtolength\height{\depth}%
  \addtolength\height{2pt}%
  \addtolength\depth{0pt}%
  {#1}_{\hspace{0.5pt}\rule[-\depth]{0.5pt}{\height} \hspace{0.5pt} {#2} }%
  }}


\newtheorem{definition}{Definition}
\newtheorem{notation}{Notation}
\newtheorem{lemma}{Lemma}
\newtheorem{proposition}{Proposition}
\newtheorem{theorem}{Theorem}
\newtheorem{corollary}{Corollary}
\newtheorem{example}{Example}
\newtheorem{constraint}{Constraint}
\newtheorem{convention}{Convention}
\newtheorem{remark}{Remark}

\newcommand{\Introduction}{\section*{Introduction}
                           \addcontentsline{toc}{chapter}{Introduction}}

\def\sl#1{\underline{#1}}
\def\cl#1{{\mathcal{#1}}}

\newcommand{\df}[1]{#1\!\!\searrow}
\newcommand{\udf}[1]{#1\!\!\nearrow}

\newcommand{\rmq}{\hspace{-5mm}{\bf Remark:} ~}
\newcommand{\lpar}{\par\noindent}


\newenvironment{infrule}{\begin{array}{c}}{\end{array}}

\def\rulename#1{({\sc #1})}

\def\et{\hskip 1.5em \relax}

\def\nm{\vspace{-1mm} \\  \hspace{-0.6cm}}
\def\nom{\vspace{-1mm} \\  \hspace{-1.1cm}}
\def\nomq{\vspace{-1mm} \\  \hspace{-1.3cm}}
\def\nomter{\vspace{-1mm} \\  \hspace{-1.4cm}}
\def\sansnom{\vspace{-2mm} \\ \hspace{-0.2cm}}
\def\nombis{\vspace{-1mm} \\  \hspace{-1.8cm}}
\def\decal{\vspace{-2mm} \\ \hspace{0.4cm}}

\def\imp{ ~ \hrulefill \\}

\def\ligne{\\[5mm]}

\def\Ligne{\\[7mm]}


\newenvironment{ex.}{
        \medskip
        \noindent {\bf Example}}
        {\hfill$\Box$ \medskip}
\newenvironment{Enonce}[1]{
        \begin{description}
        \item[{\bf #1~:}]
        \mbox{}
}{
        \end{description}
}

\def\Doubleunion#1#2#3{\displaystyle{\bigcup_{#1}^{#2} {#3}}}
\def\Union#1#2{\displaystyle{\bigcup_{#1} #2}}
\def\Intersect#1#2{\displaystyle{\bigcap_{#1} #2}}
\def\Coprod#1#2{\displaystyle{\coprod_{#1} #2}}
\def\Produit#1#2#3{\displaystyle{\prod_{#1}^{#2} {#3}}}
\def\Conj#1#2{\displaystyle{\bigwedge_{#1} \!\!\!\!\!#2}}
\def\Disj#1#2{\displaystyle{\bigvee_{#1} \!\!\!\!\!#2}}

\newcommand{\Nat}{I\!\!N}
\newcommand{\Int}{Z\!\!\! Z}

\def\infer#1#2{\mbox{\large  ${#1} \frac {#2}$ \normalsize}}

\def\sem#1{[\![ #1 ] \!]}


\newcommand{\mv}{\varepsilon}

\def\M#1#2{M_{\scriptstyle #1 \atop \scriptstyle #2}}
\def\H#1#2{H_{\scriptstyle #1 \atop \scriptstyle #2}}
\newcommand{\Ro}{{\cal{R}}}
\newcommand{\Mo}{{\cal{M}}}
\newcommand{\Ho}{{\cal{H}}}
\newcommand{\Eo}{{\cal{E}}}
\newcommand{\ssucc}{\succ \!\! \succ}


\def\id#1{\underline{#1}}

\DeclareRobustCommand\sequent{\mathrel{|}\joinrel\mkern-.5mu\mathrel{-}}

\newcommand{\ssequent}{\vdash \!\! \dashv}

\newcommand{\Function}[6][\longrightarrow]
{%
	\ensuremath{%
		\ifthenelse{\equal{#2}{}}{}{#2 : \left\lbrace }%
		\begin{array}{ccc}%
			#3 	&	 #1	& 	#4 		\\%
			#5 	& 	\longmapsto 		& 	#6%
		\end{array}%
		\ifthenelse{\equal{#2}{}}{}{\right.}%
	}%
}

\newcommand{\Functor}[8][cccc]
{
	\ensuremath{
		\ifthenelse{\equal{#2}{}}{}{#2 : \left\{ }
		\begin{array}{#1}
			#3 	&	\longrightarrow 	& 	#4 		\\
			#5 	& 	\longmapsto 		& 	#6		\\
			#7	&	\longmapsto			&	#8
		\end{array}
		\ifthenelse{\equal{#2}{}}{}{\right.}
	}
}


\def\expl#1#2{{#2}\rhd{#1}}

\newcommand{\LLE}{\mbox{\small\sf LLE}}
\newcommand{\LLEs}{\mbox{\sf LLE$_{\scriptscriptstyle\Sigma}$}}
\newcommand{\RW}{\mbox{\small\sf RW}}
\newcommand{\REF}{\mbox{\sf REF}}
\newcommand{\OR}{\mbox{\sf OR}}
\newcommand{\CM}{\mbox{\sf CM}}
\newcommand{\RM}{\mbox{\sf RM}}
\newcommand{\DR}{\mbox{\sf DR}}
\newcommand{\RT}{\mbox{\sf RT}}

\def\nms{\vdash_{\scriptscriptstyle{\Sigma}}}
\def\RLE{\mbox{\small\sf RLE}}
\def\RLEs{\mbox{\sf RLE$_{\scriptscriptstyle\Sigma}$}}
\def\con{\mbox{\sf Con$\boldmath {}_{\scriptscriptstyle\Sigma}$}}
\def\SRLE{\mbox{\sf RLE$_\Sigma$}}
\def\ECOMP{\mbox{\sf E-Comp}}
\def\Econ{\mbox{\small\sf E-Con}}
\def\ECM{\mbox{\small\sf E-CM}}
\def\EWCM{\mbox{\sf E-W-CM}}
\def\EDR{\mbox{\small\sf E-DR}}
\def\EWDR{\mbox{\small\sf LOR}}
\def\WDR{\mbox{\sf W-DR}}
\def\RA{\mbox{\small\sf RS}}
\def\ECC{\mbox{\small\sf E-C-Cut}}
\def\EWCC{\mbox{\sf E-W-C-Cut}}
\def\ERC{\mbox{\small\sf E-R-Cut}}
\def\EWRC{\mbox{\sf E-W-R-Cut}}
\def\Ecut{\mbox{\sf E-Cut}}
\def\Edisj{\mbox{\sf E-Disj}}

\def\ERW{\mbox{\small\sf ROR}}
\def\ROR{\mbox{\small\sf ROR}}
\def\LOR{\mbox{\small\sf LOR}}
\def\ERef{\mbox{\small\sf E-Reflexivity}}


\begin{frontmatter}

\title{Morpho-logic from a Topos Perspective: Application to symbolic AI}

\author{Marc Aiguier$^1$, Isabelle Bloch$^2$, Salim Nibouche$^1$ and Ram\'on Pino P\'erez$^3$}
\address{1. MICS, CentraleSupelec, Universit\'e Paris Saclay, France \\ {\it marc.aiguier@centralesupelec.fr} \\
2. Sorbonne Universit\'e, CNRS, LIP6, Paris, France \\ {\it isabelle.bloch@sorbonne-universite.fr} \\
3. CRIL-CNRS, Universit\'e d'Artois, France \\ {\it pinoperez@cril.fr }}






\begin{abstract}
Modal logics have proved useful for many reasoning tasks in symbolic artificial intelligence (AI), such as belief revision, spatial reasoning, among others. On the other hand, mathematical morphology (MM) is a theory for non-linear analysis of structures, that was widely developed and applied in image analysis. Its mathematical bases rely on algebra, complete lattices, topology. Strong links have been established between MM and mathematical logics, mostly modal logics. In this paper, we propose to further develop and generalize this link between mathematical morphology and modal logic from a topos perspective, i.e. categorial structures generalizing space, and connecting logics, sets and topology. Furthermore, we rely on the internal language and logic of topos. We define structuring elements, dilations and erosions as morphisms. Then we introduce the notion of structuring neighborhoods, and show that the dilations and erosions based on them lead to a {constructive} modal logic, for which a sound and complete proof system is proposed. We then show that the
modal logic thus defined (called morpho-logic here), is well adapted to define concrete and efficient operators for revision, merging, and abduction of new
knowledge, or even spatial reasoning.

\end{abstract}


\end{frontmatter}

\maketitle



\section{Introduction}


\Isa{Modal logics have proved useful for many reasoning tasks in symbolic artificial intelligence (AI), such as belief revision, spatial reasoning, among others. On the other hand, mathematical morphology (MM) is a theory for non-linear analysis of structures, that was widely developed and applied in image analysis. In its deterministic setting, its mathematical bases rely on algebra, complete lattices, topology. Strong links have been established between MM and mathematical logics, mostly modal logics.} Necessity $\Box$ and possibility $\Diamond$ modalities are then interpreted by the two basic MM operators, namely erosion and dilation~\cite{BHR07,IGMI_NajTal10,HR90}. This interpretation allows for easy formulations of non-classical reasoning, including revision, merging, abduction~\cite{MA:AIJ-18,MA:IJAR-18,BL02,Gorogiannis2008a}, and spatial reasoning~\cite{AO07,AB19,Blo02}. 

Erosion $\varepsilon$ and dilation $\delta$ are the two basic operations of MM often defined from a structuring element $B$ used to probe spatial structures to either dilate or erode them.  More formally, \Isa{in the set theoretical case,} let $E$ be a Euclidean space (often $\mathbb{R}^d$ or $\mathbb{Z}^d$ where $d$ is the space dimension).  Let $B$ (the structuring element) be a subset of $E$, and let $B_x =\{x+b \mid b \in B\}$ be its translation at a point $x \in E$. The dilation of a set $X$ by a structuring element $B$ is then defined as $\delta[B](X) = \{x \in E \mid \check{B}_x \cap X \neq \emptyset\}$, where $\check{B}$ is the symmetric of $B$ with respect to the origin of space, and the erosion of $X$ is defined as $\varepsilon[B](X) = \{x \in E \mid B_x \subseteq X\}$. The structuring element $B$ can be equivalently defined as a binary relation on $E$, i.e.
$$B: 
\left\{
\begin{array}{lll}
E &    \to      & \mathcal{P}(E) \\
x & \mapsto & B_x
\end{array}
\right.$$
where $\mathcal{P}(E)$ is the powerset of $E$. In this setting, the following properties hold\footnote{Note that, more generally in MM, algebraic erosions and dilations on lattices are actually defined as operations that commute with the infimum and supremum, respectively (i.e. intersection and union in $(\mathcal{P}(E), \subseteq)$). These more general forms of  operators do not necessarily involve structuring elements. The two first properties are then rather definitions in a more general setting.}:

\begin{itemize}
	\item  erosion commutes with intersection and preserves $E$, 
	\item dilation commutes with union and preserves the empty set, and 
	\item erosion and dilation defined from the same structuring element are dual operators with respect to complementation. 
\end{itemize}
All this means that erosion and dilation are modal operators, that is the tuple $(\mathcal{P}(E),\cap,\cup,\_^c,\emptyset,E,\varepsilon[B],\delta[B])$ is a modal algebra. Hence, under this interpretation, modal logic is a tool for talking about spatial transformation, and in this setting, modal logic has been applied efficiently to symbolic artificial intelligence~\cite{AB19,Blo02,BBPPT21}. 

Until now, this link between MM and logic has been studied in the set framework (with extensions to fuzzy sets).  Since then MM has been extended to a large family of algebraic structures such as graphs~\cite{cousty2013,CNS09,MS09,Vin89}, hypergraphs~\cite{BB13,IB:DAM-15}, simplicial complexes~\cite{DCN11}, \Isa{various logics}, etc. All these extensions proved useful for \Marc{knowledge representation and reasoning},  taking  into  account  low  level  information  (points  or  neighborhood  of points), structural information (e.g.  based on spatial relations between regions or objects), prior knowledge, semantics, etc. 

To take into account all of these extensions abstractly, we now propose to deepen this link between binary MM and modal logic from a topos perspective~\footnote{All the previous mentioned extensions can be represented by presheaves whose categories are toposes.}. \Marc{Hence, paraphrasing a remark by O. Caramello in~\cite{CL22}, every topos embodies a certain domain of reality, susceptible of becoming an object of knowledge (i.e. the idealized instantiations of this reality are the points of that topos).} Toposes 
constitute a categorical structure defined by A. Grothendieck in the early sixties~\cite{Gro57}, which generalizes \Marc{the notions of space, mathematical universe, and for what concerns us here,  knowledge representation}. The reason why the choice of toposes is natural is 
\Isa{as follows.}
 {From a philosophical point of view}, this choice remains in the spirit of toposes, as developed in~\cite{Vickers2007}, which can be seen as a generalization of geometric propositional logic to predicate logics, and establish in this way a connection between logics, sets and topology. As a categorial version of Lindenbaum's algebra, they can handle predicates by considering sorts as objects, functional symbols and terms as morphisms, predicates as subobjects. The setting we propose in this paper is based on this view of toposes, and aims at defining and using MM in any spatial structure. 

The proposed definitions and operations will then enhance the reasoning ability of MM, extending previous work on morphological modal logic~\cite{AB19,Blo02}. They will allow, among others, giving morphological semantics to modalities {with a topological flavor}~\cite{BB07} conventionally used for spatial reasoning~\cite{ABBG12}, and which could not be obtained directly from erosions and dilations;  
actually, the properties of these modalities are closer to the morphological operators of opening and closing (in their particular form of composition of erosion and dilation) due to a double quantification $\forall/\exists$ in their definition. In~\cite{GAB19}, MM has been extended to structuring elements based on a notion of neighborhood close to a {similar} topological notion. We then propose to extend this first work to the framework of toposes. 
To obtain the usual expected properties of erosion and dilation in this new framework, we will have to impose a supplementary condition on these new structuring neigbhorhoods which will be  an adaptation of the notion of  filter, standard in topology and logic, to the framework of topos. 
{From all this, we will then have an internal CS4-modal algebra according to the meaning given to this notion in~\cite{AMPR03} (i.e. an internal interior algebra for erosions and a weaker version of internal closure algebra for dilations because dilations will not distribute over upper bounds \Isa{(there are therefore not dilations in a strict sense)} \footnote{In~\cite{AMPR03}, the authors have shown that in the intuitionistic setting, this last property could be rejected, motivated by computer applications~\cite{FM97,PD01,Wij90}.}). This will then allow us to give a neighborhood semantics to constructive modal logic  from a topos perspective.} 


Related extensions can be found, on the logical side, in the context of institutions~\cite{AB19} and satisfaction systems~\cite{MA:AIJ-18,MA:IJAR-18}, thus encompassing many different logics in a federative framework. Applications to typical reasoning problems (revision, abduction, spatial reasoning) were instantiated in this framework. 

The interpretation of modal logic in toposes has already been approached by others. We can cite~\cite{AKK14,RZ96}. To obtain modal operators possessing the right properties (commutativity with the upper and lower bounds, and preservation of the maximum and minimum elements), these works consider adjoints between internal Heyting algebras which therefore have the good properties of preservation and commutation~\cite{MS13}. Here, we propose a less general but more constructive definition of modal operators by defining them from erosion and dilation.

Some preliminaries on toposes are given in Section~\ref{preliminaries}. We review some concepts, notations and terminology about toposes, more specifically about elementary toposes of Lawvere and Tierney~\cite{Law72}. One important contribution of this paper is to rely on the internal language of toposes, based on their logical account, which allows reasoning on them in a way close to reasoning on sets and functions. This is even more relevant in the scope of this paper where the algebraic setting of MM is considered. 
In Section~\ref{Morphology with structuring element} we formulate the notion of structuring element and two basic operators, dilation and erosion, in the framework of elementary toposes. 
In Section~\ref{sec:structuring neighborhoods}, we further extend the notion of structuring element to the notion of structuring neighborhood system. This notion was first introduced in~\cite{GAB19}, and is now generalized in toposes, with the aim of applying MM to logic for reasoning.
In Section~\ref{Morpho-logic}, we propose a new way of considering modalities in propositional modal logic, inspired by the interpretation of these modalities as dilation and erosion with a structuring neighborhood modeling an accessibility relation (useful for instance for spatial reasoning, among others). The proposed extension formalizes constructive modal logic via MM in toposes. This also extends the neighborhood semantics, usually considered on sets~\cite{Pacuit17}, to toposes. Syntax and semantics are defined, as well as a sound and complete proof system. \Isa{Finally, in Section~\ref{sec:applis}, we illustrate the proposed approach on typical examples in symbolic AI and knowledge representation, namely belief revision, merging, abduction, and spatial reasoning. Useful notations are summarized in Appendix.}

\section{Preliminaries}
\label{preliminaries}

This paper relies on many terms and notations from the categorical theory of elementary toposes. The notions introduced here make then use of basic notions of category theory (category, functor, natural transformation, limits, colimits, Cartesian closed) which are not recalled here, but interested readers may refer to textbooks such as~\cite{BW90,McL71}. 

\subsection{Notations}

In the whole paper, $\mathcal{C}$ denotes a generic category, $X$, $Y$, and $Z$ denote objects of $\mathcal{C}$. When $\mathcal{C}$ is Cartesian closed, we denote by $X^Y$ the exponential object of $X$ and $Y$. {The symbols $f$, $g$, and $h$ denote morphisms, and given a morphism $f : X \to Y$, we denote by $\dom(f) = X$ the domain of $f$ and by $\codom(f) = Y$ the co-domain of $f$}; $F,G,H : \mathcal{C} \to \mathcal{C}$ denote functors, and $\alpha,\beta : F \Rightarrow G$ denote natural transformations. 
Identity morphisms are denoted by $Id$, and initial and terminal objects by $\emptyset$ and $\mathbb{1}$, respectively. Finally, monomorphisms are denoted by $\rightarrowtail$, that is if $m$ is a monomorphism from $X$ into $Y$, then we denote it by $m: X \rightarrowtail Y$.

\subsection{Elementary Topos}
\label{topos:basic definition}

A topos $\mathcal{C}$ is a finitely complete Cartesian closed category with a subobject classifier $\Omega$. Having a subobject classifier means that there is a morphism out of the terminal object $true : \mathbb{1} \to \Omega$ such that for every monomorphism $m : Y \rightarrowtail X$ there is a unique morphism $\chi_m : X \to \Omega$ (called the characteristic morphism of $m$) such that the following diagram is a pullback:
$$
\xymatrix{
 Y \ar[r]^{!} \ar@{>->}[d]_m & \mathbb{1} \ar[d]^{true} \\
 X \ar[r]_{\chi_m} & \Omega
 }
$$

Let $X \in |\mathcal{C}|$ be an object, its set of subobjects is defined as:
$$\Sub(X) = \{[m] \mid \codom(m) = X \; \mbox{and} \; \mbox{$m$ is a monomorphism}\}$$
where $[m]$ is the equivalence class of $m$ according to the equivalence relation 
$$m \simeq m' \; \mbox{iff} \; \codom(m) = \codom(m') \; \mbox{and} \; \dom(m) \; \mbox{is isomorphic to} \; \dom(m')$$
\Isa{If there is no ambiguity, we may write simply $m$ instead of $[m]$ in the sequel.}

Let us define the partial order $\preceq_X$ on $\Sub(X)$ as follows: for all $f : Y \rightarrowtail X$ and $g :  Z \rightarrowtail X$
 $$[f] \preceq_X [g] \Longleftrightarrow \exists h : Y \rightarrowtail Z, f = g \circ h$$ 
 \Isa{A usual, $\succeq_X$ denotes the reverse order.}
 It is known that $\Sub(X)$ is a Heyting algebra~\cite{Johnstone02}, that is $(\Sub(X),\preceq_X)$ is a distributive bounded lattice with $[Id_X]$ and $[\emptyset \rightarrowtail X]$ as the largest and the smallest  elements, respectively, and which admits an implication $\to$ right-adjoint to the meet operation $\wedge$.

As $\mathcal{C}$ is finitely complete, subobjects give rise to the contravariant functor $\Sub : \mathcal{C}^{op} \to Pos$, \Isa{where $Pos$ is the category of posets,} which to every $X \in |\mathcal{C}|$ \Isa{(object of $\mathcal{C}$)} associates $\Sub(X)$ and to every \Isa{morphism} $f : X \to X'$ associates the mapping $\Sub(f) : \Sub(X') \to \Sub(X)$ which to every $[Y' \rightarrowtail X']$ associates $[Y \rightarrowtail X]$ making the diagram
$$
\xymatrix{
 Y \ar[r] \ar@{>->}[d] & Y' \ar@{>->}[d] \\
 X \ar[r] & X'
 }
$$
a pullback. 

Every topos has further the following properties~\cite{BW85,Johnstone02}:

\begin{itemize}
\item It has also finite colimits, and then it has an initial object $\emptyset$ and a 
{terminal} object $\mathbb{1}$ which are respectively the colimit and the limit of the empty diagram. 
 \item Every morphism $f$ can be factorized uniquely as $m_f \circ e_f$ where $e_f$ is an epimorphism and $m_f$ is a monomorphism. The codomain of $e_f$ is often denoted by $\im(f)$ and is called the {\em image of} $f$, and then $(A \stackrel{f}{\rightarrow} B) = (A \stackrel{e_f}{\rightarrow} \im(f) \stackrel{m_f}{\rightarrowtail} B)$.

\item Every object $X \in \mathcal{C}$ has a {\em power object} defined by $\Omega^X$ and denoted $PX$. As a power object, it satisfies the following adjunction property: 
$$\Hom_\mathcal{C}(X \times Y,\Omega) \simeq \Hom_\mathcal{C}(X,PY)$$

Given a morphism $f \in \Hom_\mathcal{C}(X \times Y,\Omega)$ (respectively $f \in \Hom_\mathcal{C}(X,PY)$) we denote by $f^\#$ its equivalent by the above bijection. The morphism $f^\#$ is called the {\em transpose} of $f$. Note that by construction, we have $(f^\#)^\# = f$.

In particular, the transpose of the identity $Id_{PX} : PX \to PX$ is the characteristic morphism of a subobject $\in_X \rightarrowtail X \times PX$ such that for every object $Y \in \mathcal{C}$ and every monomorphism $R \rightarrowtail X \times Y$, there exists a unique morphism $R \to\; \in_X$ making the following diagram a pullback:
 $$\xymatrixcolsep{4pc}{
\xymatrix{
 R \ar[r] \ar[d] & \in_X \ar[d] \\
 X \times Y \ar[r]_{Id_X \times \chi_R^\#} & X \times PX
 }}
$$
\end{itemize}

Toposes are sufficiently set-behaved to internalize a logic in which one may reason as if they were picking elements in sets, and accomodate internally constructive proofs, i.e. using neither the law of excluded middle nor the axiom of choice. We will use this internal language of toposes extensively in the paper. Now, in order not to overburden the presentation of the paper, we refer the reader to the details of this internal language in Appendix.

\section{Mathematical Morphology in Topos}
\label{Morphology with structuring element}

\subsection{\Isa{Definitions}}

MM based on structuring elements extends fairly simply to toposes.  Let $\mathcal{C}$ be a topos.

\begin{definition}[Structuring element]
A {\bf structuring element} is a morphism $b : X \to PX$  for $X \in \mathcal{C}$.
\end{definition}
By the topos properties, given a structuring element $b : X \to PX$, there exists a unique subobject {$r_b \colon R_b \rightarrowtail X \times X$ such that $b=\chi_{r_b}^\#$}, i.e. 
$$
\xymatrix{
 R_b \ar[r] \ar[d] & \mathbb{1} \ar[d] \\
 X \times X \ar[r]_-{\chi_{r_b}} & \Omega
 }
$$
is a pullback diagram. 

Let us denote by $\breve{b} : X \to PX$ the transpose of the morphism which classifies  the image of the morphism 
$
R_b \rightarrowtail X \times X \xrightarrow{\Delta_{X \times X}}
(X \times X) \times (X \times X) 
\xrightarrow{p_2 \times p_1}
X \times X
$, \Isa{where $\Delta$ denotes the diagonal morphism, and $p_i$ the projection on the $i$th space.} 
Its description in the internal language of the topos $\mathcal{C}$ is the following: 
$$\breve{b}(y) = \{x:X \mid y \in_X b(x)\}$$

\begin{definition}[Erosion]
Let $b : X \to PX$ be a structuring element. The {\bf erosion by} $b$ is the morphism $\varepsilon[b] : PX \to PX$ whose transpose classifies the morphism $r:R \rightarrowtail PX \times X$ (i.e. $\varepsilon[b] = \chi^\#_r$) where $R$ is the pullback of the diagram:
$$\xymatrixcolsep{3pc}
\xymatrix{
 R \ar[r] \ar[d] & \succeq_X \ar[d] \\
PX \times X  \ar[r]_-{Id \times b} & PX \times PX
 }
$$  
\end{definition}
In the internal logic of the topos $\mathcal{C}$, this is expressed as follows:
$$\varepsilon[b](Y) = \{x:X \mid b(x) \preceq_X Y\}$$


\begin{definition}[Dilation]
Let $b : X \to PX$ be a structuring element. The {\bf dilation by} $b$ is the morphism $\delta[b]: PX \to PX$ which classifies the image of the morphism $R \rightarrowtail X \times X \times PX \to PX \times X$ where the second morphism is the projection in the last and the first arguments, and $R$ is the pullback of the diagram:
$$\xymatrixcolsep{5pc}
\xymatrix{
 R \ar[r] \ar[d] & R_{\breve{b}} \times \in_X \ar[d] \\
X \times X \times PX  \ar[r]_-{Id \times \Delta_X \times Id} & X \times X \times X \times PX
 }
$$ 
\end{definition}
In the internal logic of the topos $\mathcal{C}$, this leads to:
$$\delta[b](Y) = \{x:X \mid\, \exists y. \; y \in_X\! \breve{b}(x) \;\wedge\; y \in_X\! Y\}$$

\begin{remark}
We could easily have extended erosions and dilations to any structuring element of the form $b : X \to PY$, and then defining morphisms $\varepsilon[b],\delta[b] : PY \to PX$. All the results given in Section~\ref{results standard MM} are easily adaptable to such structuring elements. \\

\end{remark}


Given a structuring object $b : X \to PX$, we have the two natural transformations $\overline{\varepsilon[b]},\overline{\delta[b]} : \Sub \Rightarrow \Sub$ given by the two commuting diagrams:

$
\xymatrix{
 Hom(\mathbb{1},PX) \ar[r]|-{\simeq} \ar[d]_{Hom(Id_\mathbb{1},\varepsilon[b])} & \Sub(X) \ar[d]^{\overline{\varepsilon[b]}_X} \\
Hom(\mathbb{1},PX) \ar[r]|-{\simeq} & \Sub(X)
 } \; \; 
\xymatrix{
 Hom(\mathbb{1},PX)  \ar[r]|-{\simeq} \ar[d]_{Hom(Id_\mathbb{1},\delta[b])} & \Sub(X) \ar[d]^{\overline{\delta[b]}_X} \\
Hom(\mathbb{1},PX)  \ar[r]|-{\simeq} & \Sub(X)
 }
$

We will see in the next section that when the structuring element $b$ satisfies the formula:
$$\forall x. x \in_X b(x)$$
then, $\overline{\delta[b]}$ is a \^Cech closure operator.

\subsection{Results}
\label{results standard MM}

We find all the results of MM in the set framework. The proofs of these different results mimic in the internal logic of the topos $\mathcal{C}$ the classical proofs that we can find in the setting of MM on sets. The reason is that these classical proofs are constructive intuitionistic proofs, i.e. they use neither the axiom of choice nor the law of excluded middle. The proofs being quite simple but very formal, we refer the reader to the details of the proofs given in Appendix, so as not to overload the presentation.

\begin{proposition}[Adjunction]
\label{adjunction}
The following formula in the internal logic is valid:
$$\forall Y.\;\forall Z.\; \delta[b](Y) \preceq_X Z \Longleftrightarrow  Y \preceq_X \varepsilon[b](Z)$$
\end{proposition}

{By Proposition~\ref{adjunction}, $(PX,\wedge,\vee,\Rightarrow,\varepsilon[b],\delta[b],\bot,\top)$ is an internal HGC-algebra, \Isa{i.e. a Heyting algebra equipped with an order-preserving Galois connection~\cite{DJK10}}, and then can be used to give a semantic to the intuitionistic propositonal logic with Galois connections (IntGC) introduced in~\cite{DJK10} from a topos perspective.}

\begin{proposition}
\label{extensivity, etc.}
Erosion and dilation are monotonic for $\preceq_X$ and preserve least upper bound and greater lower bound, respectively. Moreover, we have that: 
\begin{itemize}
    \item $\varepsilon[b](X) = X$, and
    \item $\delta[b](\emptyset) = \emptyset$.
\end{itemize}
Finally, $\varepsilon[b]$ and $\delta[b]$ are, respectively, anti-extensive and extensive for $\preceq_X$ {iff} the formula $\forall x.\; x \in_X\! b(x)$ is valid. More formally, these last two properties mean that both statements are valid:

\begin{enumerate}
\item $\vdash (\forall x.\; x \in_X\! b(x)) \Leftrightarrow (\forall Y.\; \varepsilon[b](Y) \preceq_X Y)$
\item $\vdash (\forall x.\; x \in_X\! b(x)) \Leftrightarrow (\forall Y.\; Y \preceq_X \delta[b](Y))$
\end{enumerate}
\end{proposition}


When $\mathcal{C}$ is complete (e.g. topos of presheaves), if we update our syntax to include set-indexed limits and colimits on any power object $PX$, then the preservation properties can be extended as follows: 
$$ 
\textstyle
\varepsilon[b](\bigwedge_{i \in I}Y_i) = \bigwedge_{i \in I}(\varepsilon[b](Y_i))
\quad \quad ; \quad \quad
\delta[b](\bigvee_{i \in I}Y_i) = \bigvee_{i \in I}(\delta[b](Y_i))
$$ for any index set $I$.
\Isa{This implies the monotony of $\varepsilon[b], \delta[b]$.}

\begin{proposition}
\label{duality}
$\varepsilon[b](\neg_X Y) = \neg_X \delta[\breve{b}](Y)$ and  $\delta[\breve{b}](\neg_X Y) \preceq_X \neg_X \varepsilon[b](Y)$where $\neg_X$ is the pseudo-complement of the internal Heyting algebra $PX$.
\end{proposition}

\begin{proof}
This directly results from the fact that the underlying internal logic for $\mathcal{C}$ is intuitionistic. 
\end{proof}

If $\mathcal{C}$ is a Boolean topos, we further have that  $\neg_X \varepsilon[b](Y) \preceq_X \delta[\breve{b}](\neg_X Y)$, \Isa{and hence the equality}.

As usual, the composition of erosion and dilation is not equal to the identity, but produces two other operators, called opening (defined as $\delta[b] \circ \varepsilon[b]$) and closing (defined as $\varepsilon[b] \circ \delta[b]$). Opening and closing have the following properties.

\begin{proposition}
\label{opening and closing}
$\varepsilon[b] \circ \delta[b]$ (closing) and $\delta[b] \circ \varepsilon[b]$ (opening) satisfy the following properties:
\begin{itemize}
\item \Isa{$\varepsilon[b] \circ \delta[b]$ and $\delta[b] \circ \varepsilon[b]$ are monotonous;}
\item $\varepsilon[b] \circ \delta[b]$ is extensive;
\item $\delta[b] \circ \varepsilon[b]$ is anti-extensive;
\item $\varepsilon[b] \circ \delta[b] \circ \varepsilon[b] = \varepsilon[b]$;
\item $\delta[b] \circ \varepsilon[b] \circ \delta[b] = \delta[b]$;
\item $\varepsilon[b] \circ \delta[b]$ and $\delta[b] \circ \varepsilon[b]$ are idempotent. 
\end{itemize}
\end{proposition}

\section{Structuring Neighborhoods: Internal Topology}
\label{sec:structuring neighborhoods}


In~\cite{GAB19}, MM based on sets has been generalized to a new setting called structuring neighborhood systems. Here, we propose to extend this generalization to the topos framework. Hence, we propose to generalize all the notions of Section~\ref{Morphology with structuring element} using a lax notion of structuring element, called {\em structuring neighborhood}. The motivation of this extension is mainly logical. Indeed, the motivations of the paper is to apply MM to logic, so-called morpho-logic (see Section~\ref{Morpho-logic}), which has been proved useful to model knowledge, beliefs or preferences, and to model classical reasoning methods such as revision, fusion, abduction or spatial reasoning~\cite{MA:AIJ-18,MA:IJAR-18,AB19,Blo02,IB:arXiv-18,BPU04}. 


\subsection{Structuring Neighborhood: Definitions and Results}

In the proposed framework of structuring neighborhoods, many properties of classical erosion and dilation can be recovered, but at the price of supplementary conditions on structuring neighborhoods. These supplementary conditions led to a mathematical construction which is very important in topology and logic, namely filters.

Let us consider an object $X$ in a topos $\mathcal{C}$. 

\begin{definition}[Filter]
\label{def:filter}
We formalize the notion of \textbf{filter} in $X$ by the following axioms on the variable $F:PPX$:

\begin{itemize}
    \item \textit{Closed under finite intersections:} 
    \[
    \forall A.\; \forall B.\; A \in_{PX}\! F \; \wedge \; B \in_{PX}\! F \; \Rightarrow \; A \wedge B \in_{PX}\! F
    \]
    \item \textit{Upper closed:} 
    \[
    \forall A.\; \forall B.\; A \in_{PX} \!F \; \wedge \; A \preceq_X\! B \; \Rightarrow \; B \in_{PX} \!F
    \]
    \item \textit{Non-empty:} 
    \[
    X \in_X F
    \]
    \item \textit{Strict:} 
    \[
    \forall A.\; A \in_{PX}\! F\; \Rightarrow\; (\exists x.\; x \in_X \!A)
    \]
\end{itemize}

The formulas above define a morphism $\mathcal{F} \colon PPX \to \Omega$.

\end{definition}
{Filters will allow us to internally define a topology over a given topos, based on the notion of {topological} neighborhood (see Definition~\ref{def:topological neighborhood}).}

\begin{definition}[Structuring neighborhood]
\label{def:structuring neighborhood}
A {\bf structuring neighborhood} is a {morphism} $N: X \to PPX$ which validates both formulas:
\begin{enumerate}
	\item $\forall x.\; \mathcal{F}(N(x))$, \Isa{where $\mathcal{F}$ is the morphism introduced in Definition~\ref{def:filter}}
	\item $\forall x. \forall A. A \in_{PX} N(x) \Rightarrow x \in_X A$
\end{enumerate}
\end{definition}

\begin{definition}[Erosion and dilation]
Let us consider a structuring neighborhood $N: X \to PPX$. Let us  define the morphism $\varepsilon[N]: PX \to PX$ by the formula:
\[
\forall Y. \; \varepsilon[N](Y) = \{x : X \; |\; Y \in_{PX} N(x) \}
\]
and the morphism $\delta[N] \colon PX \to PX$ by the formula:
\begin{align*}
\forall Y. \; \delta[N](Y) &= \{x : X \; |\; \exists F.\; \mathcal{F}(F) \; \wedge \; Y \in_{PX} F \; \wedge \; N(x) \preceq_{PX} F \}
\\
&= \{x : X \; |\; \forall A.\; A \in_{PX} N(x) \; \Rightarrow \; \exists y. \; y \in_X A \wedge Y \}
\end{align*}
\end{definition}

Given a structuring neighborhood $N : X \to PPX$, we also have two natural transformations $\overline{\varepsilon[N]},\overline{\delta[N]} : \Sub \Rightarrow \Sub$ given by the two commuting diagrams:

$
\xymatrix{
 Hom(\mathbb{1},PX)  \ar[r]|-{\simeq} \ar[d]_{Hom(Id_\mathbb{1},\varepsilon[b])} & \Sub(X) \ar[d]^{\overline{\varepsilon[N]}_X} \\
Hom(\mathbb{1},PX)  \ar[r]|-{\simeq} & \Sub(X)
 }
\; \; 
\xymatrix{
 Hom(\mathbb{1},PX)  \ar[r]|-{\simeq} \ar[d]_{Hom(Id_\mathbb{1},\delta[b])} & \Sub(X) \ar[d]^{\overline{\delta[N]}_X} \\
Hom(\mathbb{1},PX)  \ar[r]|-{\simeq} & \Sub(X)
 }
$

\begin{proposition}
\label{th:preservation results}
$\varepsilon[N]$ and $\delta[N]$ are monotonic. Moreover:

\begin{itemize}
    \item 
$\varepsilon[N]$ verifies:
\begin{itemize}
    \item $\forall A.\; \forall B.\; \varepsilon[N](A \wedge B) = \varepsilon[N](A) \wedge \varepsilon[N](B)$.
    \item $\varepsilon[N](X) = X$.
    \item $\forall Y. \varepsilon[N](Y) \preceq_X Y$.
\end{itemize}
\item $\delta[N]$ verifies:
\begin{itemize}
    \item $\forall A.\; \forall B.\; \delta[N](A \vee B) \succeq_X \delta[N](A) \vee \delta[N](B)$.
    \item $\delta[N](\emptyset) = \emptyset$.
    \item $\forall Y. Y \preceq_X \delta[N](Y)$.
\end{itemize}
\item $\forall Y. \; \varepsilon[N](\neg_X Y) \preceq_X \neg_X \delta[N](Y)$
\end{itemize}
\end{proposition}

\begin{proof}
\begin{itemize}
    \item 
    \textit{Monotony:}
    Considering variables $x:X$, $Y:PX$, $Z:PX$, and $F:PPX$, we have by definition of $\mathcal{F}$:
    \begin{align*}
        Y \preceq_X Z 
        & \;\vdash \;
        \Isa{(}(\mathcal{F}(F) \; \wedge \; Y \in_{PX} F) \; \Rightarrow \; Z \in_{PX} F \Isa{)}
        \\
        & \;\vdash \;
        \Isa{(} (\mathcal{F}(F) \; \wedge \; Y \in_{PX} F \; \wedge\; N(x) \preceq_{PX} F) 
        \\
        & \quad \quad \; \Rightarrow \; (\mathcal{F}(F) \; \wedge \; Z \in_{PX} F \; \wedge\; N(x) \preceq_{PX} F) \Isa{)}
    \end{align*}
    Then, by quantifying on $F$ and $x$ (and using the fact that if $\varphi \vdash_{x,\vec{x}} \psi$ then $\exists x.\, \varphi \vdash_{\vec{x}} \exists x.\, \psi$), we have:
    \[
        Y \preceq_X Z 
        \;\vdash \;
        \varepsilon[N](Y) \preceq_X \varepsilon[N](Z)
    \]
    \item \textit{Commutativity of $\varepsilon[N]$ with $\wedge$:} Let us consider variables $A:PX$ and $B:PX$. By monotony \Isa{(increasingness)} we already have $\varepsilon[N](A \wedge B) \preceq_X \varepsilon[N](A) \wedge \varepsilon[N](B)$. Let us prove the other inclusion.
    
    For a variable $x:X$, we have:
    \begin{align*}
        x \in_X \Isa{(} \varepsilon[N](A) \wedge \varepsilon[N](B) \Isa{)}
        & \; \Rightarrow \;
        A \in_{PX} N(x) \; \wedge \; B \in_{PX} N(x)
        \\
        & \; \Rightarrow \;
       \Isa{(} A \wedge B\Isa{)}\in_{PX} N(x)
        \\
        & \; \Rightarrow \;
        x \in_X \varepsilon[N](A \wedge B)
    \end{align*}
    \item \textit{Identity element:}
    For a variable $x:X$, we have:
    \begin{align*}
        x \in_X \varepsilon[N](X)
        & \; \Leftrightarrow \;
        X \in_{PX} N(x)
        \\
        & \; \Leftrightarrow \;
        x \in_X X
    \end{align*}
    \item \textit{Anti-extensivity:}
    For a variable $x:X$, we have:
    \begin{align*}
        x \in_X \varepsilon[N](Y)
        & \; \Leftrightarrow \;
        Y \in_{PX} N(x)
        \\
        & \; \Rightarrow \;
        x \in_X Y
    \end{align*}
    \item For dilation $\delta[N]$, the proofs of monotony, commutativity and preservation of lower bound are similar (monotony argument and definition of $\emptyset$). Let us prove the extensivity property: \Marc{Let $x : X$ be a variable. Let $x \in_X Y$. Let us assume that $\forall A. A \in_{PX} N(x)$. Then $x \in_X A$. And since $x \in_X Y$, we have $x \in_X A \wedge Y$. Therefore $x \in_X  \delta[N](Y)$ (by taking $y=x$ in the definition of the dilation).}
    \item \textit{Duality:}
    For variables $Y:PX$ and $x:X$, we have $\neg_X Y\wedge Y = \emptyset$ and then:
    \begin{align*}
        x \in_X \varepsilon[N](\neg_X Y)
        & \; \Rightarrow \;
        \neg_X Y \in_{PX} N(x)
        \\
        & \; \Rightarrow \;
        \neg_X Y \in_{PX} N(x)
        \;\wedge\;
        (\forall y.\; \neg (y \in_X \neg_X Y \wedge Y))
        \\
        & \; \Rightarrow \;
        \exists A.\; A \in_{PX} N(x) \; \wedge \; (\forall y.\; \neg (y \in_X A \wedge Y))
        \\
        & \; \Rightarrow \;
        \exists A.\; A \in_{PX} N(x) \; \wedge \; \neg (\exists y.\; y \in_X A \wedge Y)
        \\
        & \; \Rightarrow \;
        \exists A.\; \neg (A \in_{PX} N(x) \; \Rightarrow \; \exists y.\; y \in_X A \wedge Y)
        \\
        & \; \Rightarrow \;
        \neg(\forall A.\; A \in_{PX} N(x) \; \Rightarrow \; \exists y.\; y \in_X A \wedge Y)
        \\
        & \; \Rightarrow \;
        \neg(x \in \delta[N](Y))
        \\
        & \; \Rightarrow \;
        x \in \neg_X \delta[N](Y)
    \end{align*}
\end{itemize}
which concludes the proof.
\end{proof}

The definitions of $\varepsilon[N]$ and $\delta[N]$ have a topological flavor. However, they do not satisfy some properties required for interior and closure operators, respectively (they are not idempotent and $\delta[N]$ does not distribute over $\vee$).  In Section~\ref{sec:topological neighborhood}, we will restrict the set of structuring neighborhoods to topological neighborhoods (see Definition~\ref{def:topological neighborhood}). Whereas structuring neighborhoods are sufficient to give semantics to the modal logic IT, this new family of structuring neighborhoods will allow us to define erosions and dilations so that they will form an internal CS4-modal algebra over $PX$.

\begin{example}[From structuring element to neighborhood]
\label{ex:modal algebra}
Each structuring element $b \colon X \to PX$ provides a natural structuring neighborhood $N_b \colon X \to PPX$ defined by:
\[
\forall x. \; N_b(x) = \{Y : PX \; |\; Y \succeq_X b(x) \}
\]

\end{example}

\begin{proposition}
\label{prop:subsume}

$\varepsilon[N_b] = \varepsilon[b]$ and $\delta[N_b] = \delta[\breve{b}]$.

\end{proposition}

\begin{proof}
\begin{align*}
    x \in_X \varepsilon[N_b](Y)
    &\Leftrightarrow
    Y \in_{PX} N_b(x)
    \\
    &\Leftrightarrow
    Y \succeq_X b(x)
    \\
    &\Leftrightarrow
    x \in_X \varepsilon[b](Y)
\end{align*}
\begin{align*}
    x \in_X \delta[N_b](Y)
    &\Rightarrow 
    \exists F.\; \mathcal{F}(F) \; \wedge \; Y \in_{PX} F \; \wedge \; N_b(x) \preceq_{PX} F
    \\
    &\Rightarrow 
    \exists F.\; \mathcal{F}(F) \; \wedge \; Y \in_{PX} F \; \wedge \; b(x) \in_{PX} F
    \\
    &\Rightarrow 
    \exists F.\; \mathcal{F}(F) \; \wedge \; \Isa{(} Y \wedge b(x) \Isa{)} \in_{PX} F
    \\
    & \Rightarrow \exists y. \; y \in_X \Isa{(}Y \wedge b(x) \Isa{)}
    \\
    & \Rightarrow x \in \delta[\breve{b}](Y)
\end{align*}
\begin{align*}
    x \in \delta[\breve{b}](Y)
    &
    \Rightarrow \exists y. \; y \in_X \Isa{(} Y \wedge b(x) \Isa{)}
    \\
    &\Rightarrow 
    \exists F.\; F = \{Z : PX \; |\; Z \succeq_X \Isa{(} Y \wedge b(x) \Isa{)}\} 
    \\
    & \quad \quad \quad \wedge \; \mathcal{F}(F)  \; \wedge \; Y \in_{PX} F \; \wedge \; N_b(x) \preceq_{PX} F
    \\
    &\Rightarrow 
    x \in_X \delta[N_b](Y)
\end{align*}

In the second implication, we use the fact that if $Z \succeq_X Y \wedge b(x)$ and $\exists y. \; y \in_X Y \wedge b(x)$ then $\exists y. \; y \in_X Z$, which gives us the last axiom of $\mathcal{F}$. The other axioms are obviously satisfied.
\end{proof}

Hence, we proved that dilation and erosion based on structuring neighborhoods actually generalize those based on structuring elements, through the identification of each structuring element $b \colon X \to PX$ \Isa{such that $x \in_X b(x)$} with its corresponding structuring neighborhood $N_b \colon X \to PPX$.

\begin{proposition}
For all structuring neighborhoods $M,N \colon X \to PPX$, the following assertions are equivalent:
\begin{enumerate}
    \item \textit{Adjunction:} $\forall Y.\; \forall Z.\; \delta[N](Y) \preceq_X Z \; \Leftrightarrow \; Y \preceq_X \varepsilon[M](Z).$
    \item \textit{Dual structuring elements:} there exists a structuring element $b \colon X \to PX$ such that $M=N_b$ and $\delta[N]=\delta[N_{\breve{b}}]$.
\end{enumerate}
\end{proposition}

\begin{proof}
We already proved $2 \Rightarrow 1$. Let us prove $1 \Rightarrow 2$. 

If we have the adjunction property between $\delta[N]$ and $\varepsilon[M]$, then by considering $Y=\{x\}$ for any variable $x:X$, we have in particular:
\[
\forall x. \; \forall Z.\quad \delta[N](\{x\}) \preceq_X Z \; \Leftrightarrow \; \{x\} \preceq_X \varepsilon[M](Z)
\; \Leftrightarrow \; Z \in M(x) 
\]
Hence, we have necessarily:
\[
\forall x. \; M(x) = \{Z : PX \; |\; \delta[N](\{x\}) \preceq_X Z \} = N_b(x)
\]
where $b \colon X \to PX$ is the structuring element defined by:
\[
\forall x.\; b(x) = \delta[N](\{x\})
\]

We have therefore constructed a structuring element $b$ satisfying $M=N_b$. Let us now prove that we have necessarily $\delta[N]=\delta[N_{\breve{b}}]$.

By Proposition \ref{prop:subsume}, we have $\delta[N_{\breve{b}}] = \delta[b]$ and $\varepsilon[M]=\varepsilon[N_b] = \varepsilon[b]$. Therefore, Theorem \ref{adjunction}, combined with our adjunction hypothesis on $N$ and $M$, results in:
\begin{align*}
\forall Y.\; \forall Z.\quad \; 
\delta[N_{\breve{b}}](Y) \preceq_X Z 
&\; \Leftrightarrow \; 
\delta[b](Y) \preceq_X Z 
\\
&\; \Leftrightarrow \; 
Y \preceq_X \varepsilon[b](Z)
\\
&\; \Leftrightarrow \; 
Y \preceq_X \varepsilon[M](Z)
\\
&\; \Leftrightarrow \; 
\delta[N](Y) \preceq_X Z
\end{align*}

In particular, by considering respectively $Z=\delta[N_{\breve{b}}](Y)$ and $Z=\delta[N](Y)$:
\[
\forall Y. \; \delta[N](Y) \preceq_X \delta[N_{\breve{b}}](Y)
\quad \quad ; \quad \quad
\forall Y. \; \delta[N_{\breve{b}}](Y) \preceq_X \delta[N](Y) 
\]
which concludes the proof.
\end{proof}

\begin{remark}
When $\mathcal{C}$ is a Boolean topos, and then the assumption $\neg \neg \varphi = \varphi$ holds, we have moreover $N = N_{\breve{b}}$. Indeed, we have $\varepsilon[N](Y) = \neg_X \delta[N](\neg_X Y) = \neg_X \delta[b](\neg_X Y) = \varepsilon[\breve{b}](Y)$, i.e. $Y \in N(x) \, \Leftrightarrow \, \breve{b}(x)\preceq_X  Y$.
\end{remark}

\subsection{Topological Neighborhood}
\label{sec:topological neighborhood}

\begin{definition}[Topological neighborhood]
\label{def:topological neighborhood}
A {\bf topological neighborhood} is a structuring neighborhood $N: X \to PPX$ such that:
$$\forall x.\; \forall A. \; A \in_{PX}\! N(x) \Rightarrow
    \left(
    \begin{array}{l}
    \exists B.\; B \in_{PX} N(x) \; \wedge \\
    (\forall y.\; y \in_X\! B \, \Rightarrow \, A \in_{PX}\! N(y))
    \end{array}
    \right)$$
\end{definition}

\begin{proposition}[Interior operator]
\label{prop:interior algebra}
If $N : X \to PPX$ is a topological neighborhood, then $\varepsilon[N]$ is an internal interior operator in $PX$, i.e. it satisfies the following properties:
\begin{enumerate}
    \item $\forall Y.\; \varepsilon[N](Y) \preceq_X Y$.
    \item $\varepsilon[N](X) = X$.
    \item $\forall Y.\;\forall Z.\;\varepsilon[N](Y \wedge Z) = \varepsilon[N](Y) \wedge \varepsilon[N](Z)$.
    \item $\forall Y.\;\varepsilon[N](\varepsilon[N](Y)) = \varepsilon[N](Y)$.
\end{enumerate}
\end{proposition}

\begin{proof}
The first three properties have been proved in Proposition~\ref{th:preservation results}. Let us prove the last property. Let us consider a variable $Y:PX$. We already have $\varepsilon[N](\varepsilon[N](Y)) \preceq_X \varepsilon[N](Y)$. Conversely, for a variable $x:X$, we have:
\begin{align*}
    x \in_X \varepsilon[N](Y) &\; \Rightarrow \; Y \in_{PX} N(x) 
    \\
    &\; \Rightarrow \; \exists B.\; B \in_{PX} N(x) \; \wedge \; (\forall y.\; y\in_X B \; \Rightarrow \; Y \in_{PX} N(y))
    \\
    &\; \Rightarrow \; \exists B.\; B \in_{PX} N(x) \; \wedge \; (\forall y.\; y\in_X B \; \Rightarrow \; y \in_X \varepsilon[N](Y))
    \\
    &\; \Rightarrow \; \exists B.\; B \in_{PX} N(x) \; \wedge \; B \preceq_X \varepsilon[N](Y)
    \\
    &\; \Rightarrow \; \varepsilon[N](Y) \in_{PX} N(x)
    \\
    &\; \Rightarrow \; x \in_X \varepsilon[N](\varepsilon[N](Y))
\end{align*}
\end{proof}

\begin{proposition}
\label{prop:closure algebra}
If $N : X \to PPX$ is a topological neighborhood, then $\delta[N]$ satisfies the following properties:
\begin{enumerate}
    \item $\forall Y.\; Y \preceq_X \delta[N](Y)$.
    \item $\delta[N](\emptyset) = \emptyset$.
    \item $\forall Y.\;\delta[N](\delta[N](Y)) = \delta[N](Y)$.
\end{enumerate}
\end{proposition}

\begin{proof}
The first two properties have already been proved in Proposition~\ref{th:preservation results}. Let us prove the last one. Let us consider a variable $x : X$. Then, we have the following implications :
    \begin{equation}
    \label{eq:def delta delta}
    x \in_X \delta[N](\delta[N](Y)) \Rightarrow \Isa{(}\forall A. A \in_{PX} N(x) \Rightarrow \exists y. y \in_X \delta[N](Y) \wedge A \Isa{)}
    \end{equation}
    But
    \begin{equation}
    \label{eq:def topological neighborood}
    A \in_{PX} N(x) \Rightarrow \left(
                                \begin{array}{l}
                                \exists B. B \in_{PX} N(x) \wedge \\
                                (\forall y \in_X B \Rightarrow A \in_{PX} N(y))
                                \end{array}
                                \right)
    \end{equation}
    Moreover, as $B \in_{PX} N(x)$, we have that
    \begin{equation}
    \label{eq: consequence B}
        \exists y. y \in_X \delta[N](Y) \wedge B
    \end{equation}
    Now,
    \begin{equation}
    \label{eq: def delta}
        y \in_X \delta[N](Y) \Rightarrow \forall C. C \in_{PX} N(y) \Rightarrow \exists z. z \in_X Y \wedge C
    \end{equation}
    From Implications~\ref{eq:def topological neighborood} and \ref{eq: def delta}, we have that $A \in_{PX} N(y)$, and then 
    $$\exists z. z \in_X Y \wedge A$$
    from which we can conclude that $x \in \delta[N](Y)$, and hence $\delta[N](\delta[N](Y)) \preceq_X \delta[N](Y)$.
    
    Finally, $\delta[N](Y) \preceq_X \delta[N](\delta[N](Y))$ is a direct consequence of the first property, and then we have $\forall Y.\;\delta[N](\delta[N](Y)) = \delta[N](Y)$.
\end{proof}

Hence, for any topological neighborhood $N$, $\overline{\delta[N]}$ is a closure operator to which we can associate the two natural transformations $\eta : Id \Rightarrow \overline{\delta[N]}$ and $\mu : \overline{\delta[N]} \circ \overline{\delta[N]} \Rightarrow \overline{\delta[N]}$, and then $(\overline{\delta[N]},\eta,\mu)$ is a monad. 

{By Propositions~\ref{prop:interior algebra} and~\ref{prop:closure algebra}, as $\delta[N]$ does not distribute over $\vee$, we have that the tuple $(PX,\preceq_X,\wedge,\vee,\neg_X,\emptyset,\varepsilon[N],\delta[N])$ is an internal $CS4$-modal algebra~\cite{AMPR03}, and then is well adapted to give a semantic to the constructive modal logic $CS4$ from a topos perspective (which we will do in Section~\ref{Morpho-logic}).} Now, we are not far from obtaining a closure algebra. For this purpose, we must impose the supplementary condition:
$$\forall A.\;\forall B.\; \delta[N](A \vee B) \preceq_X \delta[N](A) \vee \delta[N](B)$$
\Isa{Note that there are actually some $\delta[N]$ for which this inequality holds (and then this condition combined with extensivity makes $\delta[N]$ commute with $\vee$)}.
When this holds, we have defined a notion of internal topology over $X \in \mathcal{C}$, and $\varepsilon[N]$ and $\delta[N]$ are interpreted as both topological notions of interior and closure, respectively.

In this context, we can define the two morphisms $open,closed : PX \to \Omega$ as the equalizers of $\varepsilon[N]$ and $Id_{PX}$, and $\delta[N]$ and $Id_{PX}$, respectively, which gives in the internal language of the topos $\mathcal{C}$ the following characterizations:

\begin{itemize}
    \item $open(Y) \Leftrightarrow \varepsilon[N](Y) = Y$
    \item $closed(Y) \Leftrightarrow \delta[N](Y) = Y$
\end{itemize}
We then recover the following result, which is standard for topological spaces:

\begin{proposition}
Adjunction holds if and only if opening is equivalent to closing, that is the following statement is valid: for every $N : X \to PPX$
$$Adjunction(\varepsilon[N],\delta[N]) \Leftrightarrow (\forall W,\; open(W) \Leftrightarrow close(W))$$
where $Adjunction(\varepsilon[N],\delta[N])$ is defined as:
$$Adjunction(\varepsilon[N],\delta[N]) \equiv (\forall Y.\; \forall Z.\; \delta[N](Y) \preceq_X Z \Leftrightarrow Y \preceq_X \varepsilon[N](Z))$$ 
\end{proposition}

\begin{proof}
$(\Rightarrow$) Let us assume adjunction.
Let us suppose that $Y$ is closed, that is $\delta[N](Y) = Y$, and then $\delta[N](Y) \preceq_X Y$. By adjunction, we then have that $Y \preceq_X \varepsilon[N](Y)$. But, we also know that $\varepsilon[N](Y) \preceq_X Y$, and then $\varepsilon[N](Y) = Y$. {Therefore $Y$ is also open.}
Applying this to the complement allows us to conclude that all opens are closed.

($\Leftarrow$) Let us suppose that $\delta[N](Y) \preceq_X Z$. By Theorem~\ref{th:preservation results}, we have that $$\varepsilon[N](\delta[N](Y)) \preceq_X \varepsilon[N](Z)$$ 
We know that $\delta[N](Y)$ is closed, and then it is open by hypothesis. We can then conclude that $\varepsilon[N](\delta[N](Y)) = \delta[N](Y)$. But, $\delta[N]$ is extensive, and then $Y \preceq_X \delta[N](Y)$, whence we can conclude that $Y \preceq_X \varepsilon[N](Z)$.  

The other direction of the implication of adjunction is proved similarly.
\end{proof}
\section{Morpho-Logic: Interpretation of Modalities in Topos}
\label{Morpho-logic}

Taking advantage of the fact that the tuple $(PX,\wedge,\vee,\neg_X,\emptyset,\varepsilon[N],\delta[N])$ is an internal modal algebra for the family of structuring neighborhoods $N$ defined in Definition~\ref{def:topological neighborhood}, we will give in this section a neighborhood semantics of the constructive modal logic from a topos perspective. 
As is customary in categorical logic, we propose an entailment system formulated as a sequent calculus for which we prove a completeness result. 

\paragraph{Syntax.} Let $PV$ be a countable set whose elements are called {\bf propositional variables} and denoted by letters $p,q\ldots$ The set $\Phi$ 
of formulas is defined by the following grammar:
$$\varphi,\psi ::= \top \;|\; \bot \;|\; p \;|\; \neg \varphi \;|\; \varphi \wedge \psi \;|\; \varphi \vee \psi \;|\; \varphi \Rightarrow \psi \;|\; \Box \varphi \;|\; \Diamond \varphi$$
where $p$ runs through $PV$. 

\paragraph{Semantics.} Let $\mathcal{C}$ be a topos. 
\Isa{In all this section, we assume that $\mathcal{C}$ is non-degenerated (i.e. $\emptyset \neq \mathbb{1}$, and then $true \neq false$)}. As structuring neighborhoods subsume structuring elements $b:X \to PX$ \Isa{such that $\forall x.x \in_X b(x)$} (cf. Example~\ref{ex:modal algebra}), we interpret the modal operators $\Box$ and $\Diamond$ respectively by the morphological operators $\overline{\varepsilon[N]}_X$ and $\overline{\delta[N]}_X$ for $N : X \to PPX$ with $X \in \mathcal{C}$. By Proposition~\ref{th:preservation results}, this means that $\overline{\varepsilon[N]}_X$ is anti-extensive, and $\overline{\delta[N]}_X$ is extensive, which implies that the Kripke schema holds. 

Given a set of propositional variables $PV$, a {\bf $PV$-model}, or simply a model,  $\mathcal{M}$ is a triple $(X,N,\nu)$ where:

\begin{itemize}
\item $X$ is an object of $\mathcal{C}$,
\item $N : X \to PPX$ is a structuring neighborhood, and
\item $\nu : PV \to \Sub(X)$ is a mapping called {\bf valuation}.
\end{itemize}
If we want to give semantics to CS4-modal logic, we restrict ourselves to  models $(X,N,\nu)$ where $N$ is a topological neighborhood. \\ We denote $Mod$ the class of $PV$-models.

The semantics of formulas in a model $\mathcal{M}$ is a mapping $\sem{\mathcal{M}}(\_) : \Phi \to \Sub(X)$ defined by structural induction on formulas as follows:

\begin{itemize}
\item $\sem{\mathcal{M}}(\top) = Id_X$,
\item $\sem{\mathcal{M}}(\bot) = \emptyset \rightarrowtail X$ ($\emptyset \rightarrowtail X$ is the least subobject of $\Sub(X)$),
\item $\sem{\mathcal{M}}(p) = \nu(p)$,
\item $\sem{\mathcal{M}}(\neg \varphi) = \sem{\mathcal{M}}(\varphi) \to \sem{\mathcal{M}}(\bot)$ 
\item $\sem{\mathcal{M}}(\varphi \wedge \psi) = \sem{\mathcal{M}}(\varphi) \wedge \sem{\mathcal{M}}(\psi)$ (the infimum of $\sem{\mathcal{M}}(\varphi)$ and $\sem{\mathcal{M}}(\psi)$), 
\item $\sem{\mathcal{M}}(\varphi \vee \psi) = \sem{\mathcal{M}}(\varphi) \vee \sem{\mathcal{M}}(\psi)$ (the supremum of $\sem{\mathcal{M}}(\varphi)$ and $\sem{\mathcal{M}}(\psi)$),
\item $\sem{\mathcal{M}}(\varphi \Rightarrow \psi) = \sem{\mathcal{M}}(\varphi) \to \sem{\mathcal{M}}(\psi)$
\item $\sem{\mathcal{M}}(\Box \varphi) =  \overline{\varepsilon[N]}_X(\sem{\mathcal{M}}(\varphi))$ 
\item $\sem{\mathcal{M}}(\Diamond \varphi) = \overline{\delta[N]}_X(\sem{\mathcal{M}}(\varphi))$ 
\end{itemize}
{We write $\mathcal{M} \models \varphi$ if and only if $\sem{\mathcal{M}}(\varphi) = Id_X$, and then for every $\iota \in \Sub(X)$, we write $\mathcal{M} \models_\iota \varphi$ to mean that $\iota \Isa{\preceq_X} \sem{\mathcal{M}}(\varphi))$ (\Isa{$\preceq_X$} is the ordering on $\Sub(X)$)}. Let us denote by $Mod(\varphi)$ the class of models $\mathcal{M}$ such that $\mathcal{M} \models \varphi$. Finally, given a set of formulas $\Gamma$ and a formula $\varphi$, we write $\Gamma \models \varphi$ to mean that for every model $\mathcal{M}$ which verifies $\mathcal{M} \models \psi$ for every formula $\psi \in \Gamma$, we have that $\mathcal{M} \models \varphi$.

\begin{lemma}
\label{implication lemma}
For any model $\mathcal{M} = (X, N, \nu)$, and any formulas $\varphi, \psi$, we have:
$$\mathcal{M} \models \varphi \Rightarrow \psi \; \mbox{iff} \; \sem{\mathcal{M}}(\varphi) \preceq_X \sem{\mathcal{M}}(\psi)$$
\end{lemma}

\begin{proof}
$\Sub(X)$ is a Heyting algebra, and then it satisfies the following property: $$\sem{\mathcal{M}}(\varphi) \preceq_X \sem{\mathcal{M}}(\psi) \Longleftrightarrow (\sem{\mathcal{M}}(\varphi)  \to \sem{\mathcal{M}}(\psi) = Id_X)$$
\end{proof}

\paragraph{A sound and complete entailment system.} As is customary in categorical logic, we propose an entailment system formulated as a sequent calculus. 

\begin{definition}[Sequent]
Given two formulas $\varphi,\psi \in \Phi$, a {\bf sequent} is an expression of the form $\varphi \vdash \psi$. It is {\bf valid} for a model $\mathcal{M} = (X,N,\nu)$, denoted $\varphi \models_\mathcal{M} \psi$, if we have that:  $\sem{\mathcal{M}}(\varphi) \preceq_X \sem{\mathcal{M}}(\psi)$.
\end{definition}

The sequent calculus will consist of inference rules enabling us to derive a sequent \Isa{(or a collection of sequents)} from a collection of others, which will be written:
\begin{prooftree}
\AxiomC{$\Gamma$}
\UnaryInfC{$\sigma$}
\end{prooftree}
to mean that the sequent $\sigma$ can be inferred by a collection of sequents $\Gamma$. A double line instead of the single line will mean
that each of the sequents can be inferred from the other.

Consider the following rules, where $\varphi \ssequent \psi$ is a shortened notation for $\varphi \vdash \psi$ and $\psi \vdash \varphi$:

\begin{itemize}
    \item {\em Identity rule:} 
    $$\varphi \vdash \varphi$$
    
    \item {\em Axioms:} 
    \begin{itemize}
    	\item {\bf Preservation.} $\Box \top \ssequent \top$, and $\Diamond \bot \ssequent \bot$
	\item {\bf Duality.} $\Box \neg \varphi \vdash \neg \Diamond \varphi$
	\item {\bf Distributivity.} $\Box(\varphi \wedge \psi) \ssequent \Box \varphi \wedge \Box \psi$ and $\Diamond \varphi \vee \Diamond \psi \vdash \Diamond(\varphi \vee \psi)$
	\item {\bf Axiome K.} $\Box (\varphi \Rightarrow \psi) \vdash \Box \varphi \Rightarrow \Box \psi$
	\item {\bf Axiom T.} $\Box \varphi \vdash \varphi$, and $\varphi \vdash \Diamond \varphi$
	\item {\bf Axiom S4.} $\Box \varphi \vdash \Box \Box \varphi$, and $\Diamond \Diamond \varphi \vdash \Diamond \varphi$ (when models are restricted to topological neighborhoods)
	\item {\bf Classical.} $\neg \neg \varphi \vdash \varphi$ (When $\mathcal{C}$ is a Boolean topos)
    \end{itemize}
        
    \item {\em Inconsistency:}
    $$\bot \vdash \psi$$
    
    \item {\em Tautology:}
    $$\varphi \vdash \top$$
    
    \item {\em Cut rule:}
    \begin{prooftree}
    \AxiomC{$\varphi \vdash \psi$}
    \AxiomC{$\psi \vdash \chi$}
    \BinaryInfC{$\varphi \vdash \chi$}
    \end{prooftree}
    
    \item {\em Rules for conjunction:}
    $$\hspace{.5cm} \varphi \wedge \psi \vdash \varphi \hspace{.5cm} \varphi \wedge \psi \vdash \psi \hspace{.5cm} \varphi \wedge \varphi \ssequent \varphi \hspace{.5cm} \varphi \wedge \psi \ssequent \psi \wedge \varphi$$   
    \begin{prooftree}
    \AxiomC{$\varphi \vdash \psi$}
    \AxiomC{$\varphi \vdash \chi$}
    \BinaryInfC{$\varphi \vdash \psi \wedge \chi$}
    \end{prooftree}
    
    \item {\em Rules for disjunction:}
    $$\varphi \vdash \varphi \vee \psi \hspace{1cm} \psi \vdash \varphi \vee \psi \hspace{1cm} \varphi \vee \psi \ssequent \psi \vee \varphi$$
    \begin{prooftree}
    \AxiomC{$\varphi \vdash \chi$}
    \AxiomC{$\psi \vdash \chi$}
    \doubleLine
    \BinaryInfC{$\varphi \vee \psi \vdash \chi$}
    \end{prooftree}
    
    \item {\em Distributivity:}
    $$\varphi \wedge (\psi \vee \chi) \ssequent (\varphi \wedge \psi) \vee (\varphi \wedge \chi)$$
    
    \item {\em Rule for implication:}
    \begin{prooftree}
    \AxiomC{$\varphi \wedge \psi \vdash \chi$}
    \doubleLine
    \UnaryInfC{$\varphi \vdash \psi \Rightarrow \chi$}
    \end{prooftree}
    
    \item {\em Rule for negation:}
    $$\neg \varphi \ssequent \varphi \Rightarrow \bot$$
    
    \item {\em Rule for modalities:} 
    \begin{prooftree}
    \AxiomC{$\varphi \vdash \psi$}
    \UnaryInfC{$\Box \varphi \vdash \Box \psi$}
    \end{prooftree}
    
    \begin{prooftree}
    \AxiomC{$\varphi \vdash \psi$}
    \UnaryInfC{$\Diamond \varphi \vdash \Diamond \psi$}
    \end{prooftree}
  \end{itemize}
  
  \begin{theorem}[Soundness]
\label{th:soundness}
If $\varphi \vdash \psi$ is a sequent which is provable, then it is valid for every model $\mathcal{M}$.
\end{theorem}

\begin{proof}
This is proved by structural induction on derivation. This is trivial in almost every case from the way in which semantics has been defined. For instance, the rules for modalities are a direct  consequence of the fact that for every model $\mathcal{M} = (X,N,\nu)$, $\overline{\varepsilon[N]}_X$ and $\overline{\delta[N]}_X$ are monotonic.
\end{proof} 

Let us show the completeness of the inference system. Our construction relies, as for Henkin's standard method~\cite{Henkin1949}, on the fact that any \Isa{consistent} set of formulas can be extended into a maximally consistent set of formulas. 

Let $\Gamma \subseteq \Phi$ be a set of formulas. For every $\varphi \in \Phi$, we write $\Gamma \vdash \varphi$ \Marc{to mean that} the sequent $\varphi \ssequent \top$ can be obtained by using the inference rules from all sequents $\psi \ssequent \top$ where $\psi \in \Gamma$. \Marc{$\Gamma$ is said consistent if $\Gamma \not\vdash \bot$ (i.e. $\bot \ssequent \top$ cannot be obtained from $\Gamma$).} 

By Zorn's lemma, there exists a maximally consistent set of formulas $\overline{\Gamma}$ which contains $\Gamma$ \Isa{when $\Gamma$ is consistent}.~\footnote{i.e. $\overline{\Gamma}$ is consistent, i.e. $\overline{\Gamma} \not\vdash \bot$, and there is no consistent set of formulas properly containing $\Gamma$  (i.e. for each formula $\varphi \in \Phi$, either $\varphi \in \overline{\Gamma}$ or $\neg \varphi \in \overline{\Gamma}$ but not both).} 


\begin{proposition}
\label{prop:existence of a model}
Let $\Delta$ be a maximally consistent set of formulas. There exists a model $\mathcal{M}_\Delta$ such that,
for every $\varphi \in \Phi$:
$$\sem{\mathcal{M}_\Delta}(\varphi) = Id_X \; \mbox{iff} \; \varphi \in \Delta$$
\end{proposition}

\begin{proof}
$\mathcal{M}_\Delta = (X,N,\nu)$ is any model such that
\begin{itemize}
	\item $X \in |\mathcal{C}|$ is an object different from the initial one \Isa{(this is always possible since we have assumed that $\mathcal{C}$ is non degenerated)}.
	\item 
	$
	\nu :  p \in PV \mapsto
	\left\{
	\begin{array}{ll}
	Id_X & \mbox{if $p \in \Delta$} \\
	\emptyset \rightarrowtail X & \mbox{otherwise}
	\end{array}
	\right.
	$
\end{itemize}
\Isa{Note that no constraint is imposed on $N$.}

\Isa{The equivalence in the proposition} is proved by structural induction on $\varphi$. The cases of basic formulas and propositional connectives are easily provable. Then, 
let us prove the property for the modality $\Box$ (the proof for the modality $\Diamond$ is substantially identical). So, let $\varphi$ be of the form $\Box \psi$. 
        
        $(\Longrightarrow)$ Let us suppose that $\sem{\mathcal{M}_\Delta}(\varphi) = Id_X$. \Marc{By anti-extensivity}, we necessarily have that $\sem{\mathcal{M}_\Delta}(\psi) = Id_X$. Hence, by the induction hypothesis, we have that $\psi \in \Delta$, i.e. $\psi \ssequent \top$ is obtained from $\Delta$, and then so is $\Box \psi \ssequent \top$ by the rule for modalities, whence we can then conclude that 
        $$\Box \psi \in \Delta$$
            
        $(\Longleftarrow)$ Let us suppose that $\Box \psi \in \Delta$. We then have that $\psi \in \Delta$. Otherwise, this means that the sequent $\psi \ssequent \bot$ can be obtained from $\Delta$ from which we can conclude by the axioms associated to modalities  that $\varphi \ssequent \bot$ which contradicts the hypothesis. So, by the induction hypothesis, we have that $\sem{\mathcal{M}_\Delta}(\psi) = Id_X$, from which we can conclude that $\sem{\mathcal{M}_\Delta}(\varphi) = Id_X$.
\end{proof}

\begin{theorem}[Completeness]
If $\varphi \vdash \psi$ is a sequent which is valid, then it is provable.
\end{theorem}

\begin{proof}
If $\varphi \vdash \psi$ is not provable, then the set $\Gamma = \{\neg(\varphi \Rightarrow \psi)\}$ is consistent. Let $\overline{\Gamma}$ be a maximally consistent set of formulas which contains $\Gamma$. By Proposition~\ref{prop:existence of a model}, there exists a model $\mathcal{M}_{\overline{\Gamma}}= (X,N,\nu)$ such that $\sem{\mathcal{M}_{\overline{\Gamma}}}(\neg(\varphi \Rightarrow \psi)) = Id_X$, and then $\varphi \not\models \psi$.
\end{proof}

\section{Application to Symbolic AI}
\label{sec:applis}

\subsection{Belief Revision}
\label{sub-revision}

Belief revision is the process that makes an agent's beliefs evolve with newly acquired knowledge. In a logical framework, agent's beliefs and knowledge are formally defined by formulas. In practice, in this setting, the problem is then characterized by the resolution of the inconsistency of a theory after the addition of a new formula. In practice, beliefs and knowledge of an agent are in finite number, and they can be represented by a formula. 

Modal logics are classically used to formalize beliefs. The reason is that they allow expressing beliefs about the beliefs of the other agents. Hence, we suppose here that agent's beliefs are formalized by modal formulas, and belief revision will be defined by an operator $\circ$ between two formulas $\varphi$ and $\psi$ which will express how to transform $\varphi$ into a formula $\varphi'$ such that $\varphi' \wedge \psi$ is consistent~\Isa{\cite{KM91}}.~\footnote{\Marc{$\varphi$ will often be the conjunction of a finite set of beliefs themselves defined by formulas~\cite{KM91}.}}

An axiomatization has imposed itself, the AGM theory~\cite{AGM85}, to describe the proper functioning of the revision operator $\circ$. AGM postulates are the following:

\begin{itemize}
    \item {\bf ($G1$)} If $\psi$ is a consistent formula, then so is $\varphi \circ \psi$
    \item {\bf ($G2$)} $Mod(\varphi \circ \psi) \subseteq Mod(\psi)$
    \item {\bf ($G3$)} If $\varphi \wedge \psi$ is consistent, then $\varphi \circ \psi = \varphi \wedge \psi$
    \item {\bf ($G4$)} If $\varphi \equiv \varphi'$ and $\psi \equiv \psi'$, then $Mod(\varphi \circ \psi) = Mod(\varphi' \circ \psi')$ ($\equiv$ means that formulas are logically equivalent)
    \item {\bf ($G4$')} If $\psi \equiv \psi'$, then $Mod(\varphi \circ \psi) = Mod(\varphi \circ \psi')$ 
    \item {\bf ($G5$)} $Mod((\varphi \circ \psi) \wedge \chi) = Mod(\varphi \circ (\psi \wedge \chi))$ if $(\varphi \circ \psi) \wedge \chi$ is consistent
\end{itemize}
\Marc{Postulate ($G4$) means a complete independence of the syntaxe. But, we will see next that when the revision operator applies a syntactic transformation to knowlege bases (here specified by the formula $\varphi$) Postulate ($G4$) cannot be ensured anymore. This will be the case for our revision operator dedicated to the logic CS4 (see below), hence the introduction of Postulate ($G4$').}

\paragraph{Revision operators based on dilations.} By the way we interpreted modalities, we have for every $\varphi \in \Phi$ that
$$Mod(\varphi) \subseteq Mod(\Diamond \varphi)$$
(This is just a consequence of the fact that dilations are extensive).

Following the approach developed in~\cite{BBPPT21}, the idea is then to dilate the class of models of $\varphi$ until meeting the class of models of $\psi$. Then, as this has been defined in~\cite{BBPPT21} for the propositional logic, let us define the operator $\circ$ as follows:
$$\varphi \circ \psi = \Diamond^n \varphi \wedge \psi$$
with $n = \min\{k \in \mathbb{N} \mid \Diamond^k \varphi \wedge \psi \; \mbox{is consistent}\}$ and $\Diamond^n \varphi = \underbrace{\Diamond \ldots \Diamond}_{n~times} \varphi$.

\begin{proposition}
\label{prop:staisfies AGM}
The operator $\circ$ satisfies the AGM postulates ($G1$)-($G5$).
\end{proposition}

\begin{proof}
$\circ$ obviously satisfies Postulates $(G1)$, $(G2)$ and $(G3)$. 

\Isa{Let us prove $(G4)$.} Let $\varphi \equiv \varphi'$ and $\psi \equiv \psi'$. This means that $Mod(\varphi) = Mod(\varphi')$ and $Mod(\psi) = Mod(\psi')$. By definition of the operator $\circ$, we have that $Mod(\varphi \circ \psi) = Mod(\Diamond^n \varphi) \cap Mod(\psi)$. Hence, we have that $Mod(\varphi \circ \psi) = Mod(\Diamond^n \varphi) \cap Mod(\psi')$. Likewise, $Mod(\varphi \circ \psi') = Mod(\Diamond^m \varphi) \cap Mod(\psi')$, and then $Mod(\varphi \circ \psi') = Mod(\Diamond^m \varphi) \cap Mod(\psi)$. Obviously, we have for every $k \in \mathbb{N}$ that $Mod(\Diamond^k \varphi) = Mod(\Diamond^k \varphi')$. This entails that both $n \leq m$ and $m \leq n$, i.e. $n = m$, \Isa{and $Mod(\varphi \circ \psi) = Mod(\varphi' \circ \psi')$}. 

\Isa{Let us now prove $(G5)$.}
As $(\varphi \circ \psi) \wedge \chi$ is consistent, we have that $Mod((\varphi \circ \psi) \wedge \chi) \neq \emptyset$. Let $\mathcal{M} \in Mod((\varphi \circ \psi) \wedge \chi)$. This means that $\mathcal{M} \in Mod(\Diamond^n \varphi \wedge \psi \wedge \chi)$. By definition of the operator $\circ$, $\varphi \circ (\psi \wedge \chi) = \Diamond^m \varphi \wedge \psi \wedge \chi$. This means that $n \leq m$, and then as $\Diamond$ is extensive, we have that $\mathcal{M} \in Mod(\Diamond^m \varphi)$, and then $\mathcal{M} \in Mod(\varphi \circ (\psi \wedge \chi))$. The opposite inclusion is proved by the same arguments.
\end{proof}

\Marc{The above approach does not work if we consider Axiom S4 \Isa{of modal logics} to be valid. Indeed, in this case, the sequent $\Diamond \Diamond \varphi \ssequent \Diamond \varphi$ is valid.} Another way to define a revision operator is then to change necessity modalities to possibility ones. Indeed, it is quite intuitive that if \Isa{the revision} cannot be consistent for all states, it can be for some of them. A similar approach has been adopted in~\cite{MA:AIJ-18} for defining revision operators in first-order logic and modal logic in a set framework. The definition of these revision operators took advantage of Boolean reasoning. Thus, the revision operators were defined on formulas in normal form. In an intuitionistic framework, such normal forms for formulas do not exist. So, when the topos $\mathcal{C}$ is not Boolean, we propose the following definition for the revision operator: let us define \Marc{first} two mappings $\rho,\kappa : \Phi \to \Phi$ on formulas as

\begin{itemize}
    \item $\rho(\top) = \top$ and $\kappa(\top) = \top$
    \item $\rho(\bot) = \bot$ and $\kappa(\bot) = \bot$
    \item $\rho(p) = p$ and $\kappa(p) = p$ with $p \in PV$.
    \item $\rho(\varphi \Rightarrow \psi) = (\kappa(\varphi) \Rightarrow \psi) \vee (\varphi \Rightarrow \rho(\psi))$ and \\ $\kappa(\varphi \Rightarrow \psi) = (\varphi \Rightarrow \kappa(\psi)) \vee (\rho(\varphi) \Rightarrow \psi)$
    \item $\rho(\varphi~@~\psi) = (\rho(\varphi)~@~\psi) \vee (\varphi~@~\rho(\psi))$ and \\ $\kappa(\varphi~@~\psi) = (\kappa(\varphi)~@~\psi) \vee (\varphi~@~\kappa(\psi))$ with $@ \in \{\wedge,\vee\}$
    \item $\rho(\neg \varphi) = \neg\kappa(\varphi)$ and $\kappa(\neg \varphi) = \neg \rho(\varphi)$
    \item $\rho(\Box \varphi) = \Diamond \varphi$ and $\kappa(\Box \varphi) = \Box \kappa(\varphi)$
    \item $\rho(\Diamond \varphi) = \Diamond \rho(\varphi)$ and $\kappa (\Diamond \varphi) = \Box \varphi$
\end{itemize}

\begin{proposition}
\label{prop:revision preserves order}
For every formula $\varphi$ and every model $\mathcal{M} = (X,N,\nu)$, we have $\sem{M}(\varphi) \preceq_X \sem{M}(\rho(\varphi))$ and $\sem{M}(\kappa(\varphi)) \preceq_X \sem{M}(\varphi)$.
\end{proposition}

\begin{proof}
The proof is done by structural induction on $\varphi$. The basic cases are obvious. For the general case, several forms of $\varphi$ must be considered. Here, we prove the result for implication and negation. The other cases are quite direct.

\begin{itemize}
    \item $\varphi$ is of the form $\psi \Rightarrow \chi$.
    \begin{itemize}
        \item For the mapping $\rho$. By definition, we have that $$\sem{\mathcal{M}}(\psi \Rightarrow \chi) = \sem{\mathcal{M}}(\psi) \to \sem{\mathcal{M}}(\chi)$$ 
        By definition of $\to$, $\sem{\mathcal{M}}(\psi) \to \sem{\mathcal{M}}(\chi)$ is the greatest element $\iota$ in $\Sub(X)$ such that
        $$\sem{\mathcal{M}}(\psi) \wedge \iota \preceq_X \sem{\mathcal{M}}(\chi)$$
        By the induction hypothesis, we have both $\sem{\mathcal{M}}(\kappa(\psi)) \preceq_X \sem{\mathcal{M}}(\psi)$ and $\sem{\mathcal{M}}(\chi) \preceq_X \sem{\mathcal{M}}(\rho(\chi))$. Thus, we have that
        $$\sem{\mathcal{M}}(\kappa(\varphi)) \wedge \sem{\mathcal{M}}(\psi) \to \sem{\mathcal{M}}(\chi) \preceq_X \sem{\mathcal{M}}(\chi)$$
        and
        $$\sem{\mathcal{M}}(\varphi) \wedge \sem{\mathcal{M}}(\psi) \to \sem{\mathcal{M}}(\chi) \preceq_X \sem{\mathcal{M}}(\rho(\chi))$$
        and then by the universal property of $\to$ \Isa{(i.e. adjoint of $\wedge$)}, we have that $$\sem{\mathcal{M}}(\psi \Rightarrow \chi) = \sem{\mathcal{M}}(\rho(\varphi \Rightarrow \chi))$$
        \item For the mapping $\kappa$. By definition, we have that $$\sem{\mathcal{M}}(\psi \Rightarrow \kappa(\chi)) = \sem{\mathcal{M}}(\psi) \to \sem{\mathcal{M}}(\kappa(\chi))$$
        and
        $$\sem{\mathcal{M}}(\rho(\psi) \Rightarrow \chi) = \sem{\mathcal{M}}(\rho(\psi)) \to \sem{\mathcal{M}}(\chi)$$
        By definition of $\to$, $\sem{\mathcal{M}}(\psi) \to \sem{\mathcal{M}}(\kappa(\chi))$ and $\sem{\mathcal{M}}(\rho(\psi)) \to \sem{\mathcal{M}}(\chi)$ are the greatest elements $\iota$ and $\theta$ of $\Sub(X)$ such that 
        $$\sem{\mathcal{M}}(\psi) \wedge \iota \preceq_X \sem{\mathcal{M}}(\kappa(\chi))$$
        and 
        $$\sem{\mathcal{M}}(\rho(\psi)) \wedge \theta \preceq_X \sem{\mathcal{M}}(\chi)$$
        By the induction hypothesis, we have that $\sem{\mathcal{M}}(\kappa(\chi)) \preceq_X \sem{\mathcal{M}}(\chi)$ and $\sem{\mathcal{M}}(\psi) \preceq_X \sem{\mathcal{M}}(\rho(\psi))$, and then by the universal property of $\to$, we have that 
        $$\sem{\mathcal{M}}(\psi \Rightarrow \kappa(\chi)) \preceq_X \sem{\mathcal{M}}(\psi \Rightarrow \chi)$$
        and
        $$\sem{\mathcal{M}}(\rho(\psi) \Rightarrow \chi) \preceq_X \sem{\mathcal{M}}(\psi \Rightarrow \chi)$$
    \end{itemize}
    \item $\varphi$ is of the form $\neg \psi$.
    \begin{itemize}
        \item For the mapping $\rho$. By definition, we have that $$\sem{\mathcal{M}}(\neg \psi) = \sem{\mathcal{M}}(\psi) \to \sem{\mathcal{M}}(\bot)$$ 
        By definition of $\to$, we have 
        $$\sem{\mathcal{M}}(\psi) \wedge \sem{\mathcal{M}}(\neg \psi) \preceq_X \sem{\mathcal{M}}(\bot)$$
        By the induction hypothesis, we have that $\sem{\mathcal{M}}(\kappa(\psi)) \preceq_X \sem{\mathcal{M}}(\psi)$, and then by the universal property of $\to$, we can conclude that
        $$\sem{\mathcal{M}}(\neg \psi) \preceq_X \sem{\mathcal{M}}(\neg \kappa(\psi)\Isa{)} $$
        \Isa{and therefore $\sem{\mathcal{M}}(\varphi) \preceq_X \sem{\mathcal{M}}(\rho(\varphi))$}.
        \item For the mapping $\kappa$. By definition, we have that
        $$\sem{\mathcal{M}}(\kappa(\neg \psi)) \preceq_X \sem{\mathcal{M}}(\rho(\psi)) \to \sem{\mathcal{M}}(\bot)$$
        
        By definition of $\to$, we then have that
        $$\sem{\mathcal{M}}(\rho(\psi)) \wedge \sem{\mathcal{M}}(\kappa(\neg(\psi)) \preceq_X \sem{\mathcal{M}}(\bot)$$
        By the induction hypothesis, we have that $\sem{\mathcal{M}}(\kappa(\neg \psi)) \preceq_X \sem{\mathcal{M}}(\neg \psi)$, and then by the universal property of $\to$, we can conclude that 
        $$\sem{\mathcal{M}}(\kappa(\neg \psi)) \preceq_X \sem{\mathcal{M}}(\neg \psi)$$
    \end{itemize}
\end{itemize}
\end{proof}

Let us define the mapping $\tau : \Phi \to \Phi$ as follows:
$$
\tau : 
\left\{
\begin{array}{lll}
   \Phi  & \to & \Phi  \\
    \varphi & \mapsto & \left\{
                        \begin{array}{ll}
                          \chi   & \mbox{if $\rho(\varphi) = \varphi$}  \\
                           \rho(\varphi)  & \mbox{otherwise} 
                        \end{array}
                        \right.
\end{array}
\right.
$$
\Isa{where $\chi$ is a tautology.}

\begin{proposition}
For every formula $\varphi$, we have:
\begin{itemize}
    \item {\bf Extensivity.} $Mod(\varphi)  \subseteq Mod(\tau(\varphi))$
    \item {\bf Exhaustivity.} There exists $k \in \mathbb{N}$ such that $Mod(\tau^k(\varphi)) = Mod$
\end{itemize}
\end{proposition}

\Isa{The operator $\tau$ is then an extensive operator that is not a dilation.}

\begin{proof}
Obviously, Exhaustivity holds. Let $\mathcal{M} \in Mod(\varphi)$. By definition of semantics, this means that $\sem{\mathcal{M}}(\varphi) = Id_X$. If $\varphi$ is a propositional formula, we know that $Mod(\varphi) \subseteq Mod(\Diamond \varphi)$. Otherwise, by Proposition~\ref{prop:revision preserves order}, we have that $\sem{\mathcal{M}}(\varphi) \preceq_X \sem{\mathcal{M}}(\rho(\varphi))$, and then $\sem{\mathcal{M}}(\rho(\varphi)) = Id_X$.
\end{proof}

Let us define the following revision operator $\circ$:
$$\varphi \circ \psi = \tau^n(\varphi) \wedge \psi$$
with $n = \min\{k \in \mathbb{N} \mid \tau^k(\varphi) \wedge \psi \; \mbox{is consistent}\}$ and $\tau^n(\varphi) = \underbrace{\tau( \ldots \tau}_{n~times} (\varphi) \ldots )$.

\begin{proposition}\label{op-revision2} 
The operator $\circ$ satisfies the AGM postulates ($G1$)-($G3$), ($G4$'), and ($G5$).
\end{proposition}

\begin{proof}
The proof is substantially similar to that of Proposition~\ref{prop:staisfies AGM}.
\end{proof}

\paragraph{Minimality result.} In a previous paper~\cite{MA:AIJ-18}, we showed independently of any logic a minimization result demonstrated first by H. Katsuno and A.-O. Mendelzon~\cite{KM91} in the framework of propositional logic. This result is a representation theorem which means that revision operators satisfying AGM postulates induce minimal changes, that is the models of $\varphi \circ \psi$ are the models of \Isa{$\psi$ that are the closest to models of $\varphi$} according to some distance for measuring how close are
models. This result is based on the notion of faithful assignment whose definition we recall below. 

\begin{definition}[Faithful assignment] An {\bf assignment} is a mapping that assigns to each formula $\varphi$ a binary relation $\preceq_\varphi$ on models. We say that this assignment is {\bf faithful (FA)} if the following two conditions are satisfied:

\begin{enumerate}
    \item If $\mathcal{M},\mathcal{M}' \in Mod(\varphi)$, then $\mathcal{M} \not\prec_\varphi \mathcal{M}'$
    \item For every $\mathcal{M} \in Mod(\varphi)$ and for every model $\mathcal{M} \notin Mod(\varphi)$, $\mathcal{M} \prec_\varphi \mathcal{M}'$.
\end{enumerate}
($\mathcal{M} \prec_\varphi \mathcal{M}'$ means that $\mathcal{M} \preceq_\varphi \mathcal{M}'$ and $\mathcal{M}' \not\preceq_\varphi \mathcal{M}$).
\end{definition}

\begin{proposition}
\label{prop:minimality}
A revision operator $\circ$ satisfies the AGM postulates if and only if there exists a FA that maps each formula $\varphi$ to a binary relation $\preceq_\varphi$ such that for every formula $\psi$ 
$$Mod(\varphi \circ \psi) = Min(Mod(\psi),\preceq_\varphi)$$
where $Min(Mod(\psi),\preceq_\varphi) = \{\mathcal{M} \in Mod(\psi) \mid \forall \mathcal{M}' \in Mod(\psi), \mathcal{M}' \not\prec_\varphi \mathcal{M}\}$.
\end{proposition}

The proof of Proposition~\ref{prop:minimality} is given in Appendix. It is substantially similar to that of the general result given in~\cite{MA:AIJ-18} (see Theorem 1) except that the result here relates to formulas and not to theories. This then required redefining a particular FA in the proof. The FA used in~\cite{MA:AIJ-18} to prove this representation result was an adaptation of the FA proposed in the original paper~\cite{KM91}, but the latter was not adaptable within the framework of modal logic~\footnote{The reason is that, unlike in propositional logic, the set of valid modal formulas for a model is not necessarily definable by a single modal formula.}.

\subsection{\Ramon{Contraction}}
\Ramon{Other important operators considered in the domain of belief change are contraction operators~\cite{AGM85}. Unlike revision operators, the goal of  contraction operators is to remove the new piece of information.}

\Ramon{Using the results of the previous subsection in which revision operators have been defined, we can now define, via the  Harper identity~\cite{Harper1976},  contraction operators in the following way:  }
\Ramon{
\[\varphi \contrac \psi= (\varphi \circ \neg \psi)\vee \varphi   \quad\quad  \mbox{\bf (Harper identity)} \]
}
\Ramon{The well behaved contraction operators should satisfy the following postulates:\footnote{\Ramon{We present the formulation of postulates as in \cite{CKM17}.  }}}

\Ramon{\noindent \textbf{(C1)} $ \varphi\vdash \varphi\contrac\psi $\\[1.2mm]
\textbf{(C2)} If $ \varphi\nvdash\psi $, then $ \varphi\contrac\psi\vdash \varphi $\\[1.2mm]
\textbf{(C3)} If $ \varphi\contrac\psi\vdash\psi $, then $ \vdash\psi $\\[1.2mm]
\textbf{(C4)} If $ \varphi\vdash\psi $, then $ (\varphi\contrac\psi)\wedge\psi\vdash \varphi $\\[1.2mm]
\textbf{(C5)} If  $\psi_{1}\equiv\psi_{2} $
 then $ \varphi\contrac\psi_{1}\equiv \varphi\contrac\psi_{2} $\\[1.2mm]
\textbf{(C6)} $ \varphi\contrac(\psi\wedge\beta)\vdash (\varphi\contrac\psi)\vee  (\varphi\contrac\beta) $\\[1.2mm]
\textbf{(C7)} If $ \varphi\contrac(\psi\wedge\beta)\nvdash\psi $,
then $\varphi\contrac\psi\vdash \varphi\contrac(\psi\wedge\beta) $}

\Ramon{Now we can establish the following result, the proof of which is a straightforward consequence of Proposition~\ref{op-revision2} and standard facts about the duality between revision and contraction~\cite{AGM85}: }

\Ramon{
\begin{proposition}
   Let $\contrac$ be the operator defined by the Harper identity where $\circ$ is the revision operator of Proposition~\ref{op-revision2}. Then,  $\contrac$ satisfies (C1-C7).
\end{proposition}
}

\subsection{\Isa{Merging}}

\Isa{When the beliefs play symmetrical roles, another widely addressed problem is the one of belief merging, or fusion. Let $\varphi_1... \varphi_m$ be $m$ formulas modeling agents' beliefs. As for revision, their fusion can be defined from a dilation $\delta$ or better by a $\tau$ operator (defined as in Section~\ref{sub-revision}) as:
$$Merging(\varphi_1... \varphi_m) = \tau^n(\varphi_1) \wedge ... \wedge \tau^n(\varphi_m)$$
where $\tau$ is defined as for the revision, and
$$n = \min \{k \in \mathbb{N} \mid \wedge_{i=1}^m \tau^k(\varphi_i) \mbox{ is consistent} \}$$
It has been shown in~\cite{IB:arXiv-18,BPU04} that this dilation-based merging is equivalent to a merging operator defined from a specific distance and aggregation function, and satisfies the set of rationality postulates introduced in~\cite{KP11}, except the independence of the syntax, since syntactic operations are involved in $\tau$.~\footnote{\Marc{Of course, if we define the fusion by $$Merging(\varphi_1... \varphi_m) = \Diamond^n(\varphi_1) \wedge ... \wedge \Diamond^n(\varphi_m)$$ then all the rationality postulates in~\cite{KP11} are satisfied because the complete independence of the syntax holds.}}
}

\subsection{Abduction}

Abduction is the process of finding, given a theory $T$ and an observation $\varphi$, the best explanation $\psi$ such that $T \cup \{\psi\} \models \varphi$. Candidate explanations of $\varphi$ with respect to $T$ are many. In a logical framework, following our previous work in~\cite{MA:IJAR-18} where abduction has been studied independently of a given logic, we will also study abduction as a form of inference. Intuitively, finding candidate explanations for $\varphi$ with respect to $T$ consists in cutting/eroding the class of models of $T$ 
\Isa{while remaining consistent with} $\varphi$. Hence, abduction can be seen as the dual of revision, and then it will consist in cutting in $Mod(T \cup \{\varphi\})$ as much as possible while preserving minimality properties. For this, we propose to instantiate the abstract approach developed in~\cite{MA:IJAR-18} to our framework of categorical morpho-logic~\footnote{The abstract approach in~\cite{MA:IJAR-18} has been applied to the case of modal logic but in the set-theoretic framework. Here, as before, we must adapt this instantiation to the intutionistic framework.}. In an intuitionistic framework, to work around the problem of missing normal forms of formulas, we propose the mapping $\zeta: \Phi \to \Phi$ defined as follows: let $\chi$ be an antilogy
$$
\zeta : 
\left\{
\begin{array}{lll}
\Phi & \to & \Phi \\
\varphi & \mapsto & \left\{
                    \begin{array}{ll}
                    \chi & \mbox{if $\kappa(\varphi) = \varphi$} \\
                    \kappa(\varphi) & \mbox{otherwise}
                    \end{array}
                    \right.
\end{array}
\right.
$$
where $\kappa$ is the mapping defined in Section~\ref{sub-revision}.

\begin{proposition}
The mapping $\zeta$ verifies for every consistent formula $\varphi \in \Phi$:

\begin{itemize}
    \item {\bf Anti-extensivity.} $Mod(\zeta(\varphi)) \subseteq Mod(\varphi)$
    \item {\bf Vacuum.} There exists $k \in \mathbb{N}$ such that $Mod(\zeta^k(\varphi)) = \emptyset$.
\end{itemize}
\end{proposition}

In~\cite{MA:IJAR-18}, mappings satisfying such conditions are called retractions.

\begin{proof}
$\zeta$ is anti-extensive because by Proposition~\ref{prop:revision preserves order} $\kappa$ is. The vacuum property holds because in a finite number of steps, we always reach the antilogy $\chi$.
\end{proof}

Following~\cite{MA:IJAR-18}, from $\zeta$ we can define two families of model classes $\mathcal{C}_{lcr}$ and $\mathcal{C}_{lnr}$ as follows~\footnote{$lcr$ for {\em last consistent retraction} and $lnr$ for {\em last non-trivial retraction}}: Let $T$ be a finite set of modal formulas and let $\varphi$ be a formula such that $T \cup \{\varphi\}$ is consistent
$$\mathcal{C}^\varphi_{lcr} = \{Mod(\zeta^k(\bigwedge T) \wedge \varphi) \mid k \in \mathbb{N}, Mod(\zeta^k(\bigwedge T) \wedge \varphi) \neq \emptyset\}$$
$$\mathcal{C}^\varphi_{lnr} = \{Mod(\zeta^k(\bigwedge T \wedge \varphi)) \mid k \in \mathbb{N}, Mod(\zeta^k(\bigwedge T \wedge \varphi)) \neq \emptyset\}$$
where $\bigwedge T = \varphi_1 \wedge \ldots \wedge \varphi_n$ if $T = \{\varphi_1,\ldots,\varphi_n\}$.

Let us observe that in both $\mathcal{C}^\varphi_{lcr}$ and $\mathcal{C}^\varphi_{lnr}$ we have a unique maximal chain of finite size due to the vacuum property. Hence, both $\mathcal{C}^\varphi_{lcr}$ and $\mathcal{C}^\varphi_{lnr}$ are closed under set-theoretical inclusion and are well-founded. In~\cite{MA:IJAR-18}, any sub-family satisfying such conditions is called cutting. 

These two cuttings give rise to the following two explanatory relations:
$$\varphi \rhd_{\mathcal{C}_{lcr}} \psi \Longleftrightarrow 
\left\{
\begin{array}{l}
Mod(T \cup \{\psi\}) \neq \emptyset,~\mbox{and} \\
Mod(T \cup \{\psi\}) \subseteq Mod(\zeta^n(\bigwedge T) \wedge \varphi)
\end{array}
\right.$$
where $n = \sup\{k \in \mathbb{N} \mid Mod(\zeta^k(\bigwedge T) \wedge \varphi) \neq \emptyset\}$;
$$\varphi \rhd_{\mathcal{C}_{lnr}} \psi \Longleftrightarrow 
\left\{
\begin{array}{l}
Mod(T \cup \{\psi\}) \neq \emptyset,~\mbox{and} \\
Mod(T \cup \{\psi\}) \subseteq Mod(\zeta^m(\bigwedge T \wedge \varphi))
\end{array}
\right.$$
where $m = \sup\{k \in \mathbb{N} \mid Mod(\zeta^k(\bigwedge T \wedge \varphi)) \neq \emptyset\}$.

Directly from Theorems 2, 3 and 4 in~\cite{MA:IJAR-18}, we derive that these two explanatory relations satisfy all or part of the (rationality) postulates defined in~\cite{PPU99} which we summarize in Table~\ref{tab:summary}.

\begin{table}[htbp]
\begin{center}
\begin{tabular}{lcc} \hline
Rationality postulates & $\rhd_{\mathcal{C}_{lcr}}$ & $\rhd_{\mathcal{C}_{lnr}}$ \\ \hline \hline
\LLE\ and \RLE & $\surd$ & $\surd$ \\
\RA & $\surd$ & $\surd$ \\
\Econ & $\surd$ & $\surd$ \\ \hline
\ROR & $\surd$ & $\surd$ \\
\ERef & $\surd$ & $\surd$ \\ \hline
\ECM & $\surd$ &  \\
\ECC & $\surd$ &  \\ \hline
\end{tabular}
\end{center}
\caption{Links between rationality postulates in~\cite{PPU99} and properties satisfied by $\rhd_{\mathcal{C}_{lcr}}$ and $\rhd_{\mathcal{C}_{lnr}}$.\label{tab:summary}}
\end{table}

The original postulates of~\cite{PPU99} are recalled in Table~\ref{tab:postulates}. 

\begin{table}[htbp]
\begin{tabular}{lcl}
\LLE: & & If $\nms \alpha\leftrightarrow \alpha'$ and $\expl{\gamma}{\alpha}$ then $\expl{\gamma}{\alpha'}$.\\

\RLE: & & If $\nms \gamma \leftrightarrow \gamma'$ and $\expl{\gamma}{\alpha}$ then $\expl{\gamma' }{\alpha}$.\\

\ECM: & & If $\expl{\gamma}{\alpha} $ and $\gamma\nms\beta $ then $ \expl{\gamma}{(\alpha\wedge\beta)}$.\\

\ECC:   & & If $ \expl{\gamma}{(\alpha\wedge \beta) }$ and $\forall \delta \;[\expl{\delta}{\alpha} \;\Rightarrow \;
\delta\nms \beta\;] $ then $\expl{\gamma}{\alpha} $.\\

\RA:    & & If $ \expl{\gamma}{\alpha}$, $\gamma'\nms\gamma$ and $\gamma'\not\nms\bot $ then $ \expl{\gamma'}{\alpha}$.\\

\ERW:      && If $ \expl{\gamma}{\alpha}$ and $\expl{\delta}{\alpha} $ then $\expl{(\gamma\vee\delta)}{\alpha} $.\\

%
%

\ERef:       && If $\expl{\gamma}{\alpha} $  then $\expl{\gamma}{\gamma} $.\\

\Econ:      &&  $ \not\nms \neg \alpha$ iff there is $\gamma$ such that $\expl{\gamma}{\alpha}$.
\end{tabular}
\caption{Rationality postulates of~\cite{PPU99}, expressed according to a theory $\Sigma$.}
\label{tab:postulates}
\end{table}

\subsection{Spatial Reasoning: RCC-8}

A part of the domain of qualitative spatial reasoning deals with topological relations between spatial entities. Here we propose to apply the defined morpho-logic to mereotopology, and more precisely to the RCC-8 model~\cite{RAND-92}. This theory allows defining several topological relations from a connection predicate $C$, and has been often formalized in first order logic in the literature.
Let us recall the eight relations of RCC-8 theory (without their formalization):

\begin{itemize}
\item {\bf Disconnection $DC$.} $DC(XY)$ means that region $X$ is disconnected from region $Y$; 
\item {\bf External Connection $EC$.} $EC(X,Y)$ means that $X$ is externally connected to $Y$ (close to the notion of adjacency);
\item {\bf Partial Overlap $PO$.} $PO(X,Y)$ means that $X$ and $Y$ intersect each other but are not equal;
\item {\bf Tangential Proper Part (resp. inverse) $TPP$ (resp. $TPP_i$).} $TPP(X,Y)$ (resp. $TPP_i(X,Y)$) means that $X$ (resp. $Y$) is a tangential proper part of $Y$ (resp. of $X$);
\item {\bf  Non-Tangential Proper Part (resp. inverse) $NTPP$ (resp. $NTPP_i$).} $NTPP(X,Y)$ (resp. $NTPP_i(X,Y)$) means that $X$ (resp. $Y$) is a non-tangential proper part of $Y$ (resp. of $X$);
\item {\bf Equality $EQ$.} $EQ(X,Y)$ means that $X$ and $Y$ are identical regions. 
\end{itemize}
As shown in~\cite{AB19,Blo02,BBPPT21}, morpho-logic, as defined in this paper, provides a simple axiomatization of some of these relations. 


Let $\mathcal{M} = (X,N,\nu)$ be a model. The sub-objects of $X$ are spatial entities (i.e. regions), and formulas are combinations of such entities. The RCC-8 relations can then be formulated as follows:

\begin{itemize}
\item $C(X,Y) : \varphi \wedge \psi$;
\item $DC(XY) : \neg (\varphi \wedge \psi)$; 
\item $EC(X,Y) :$ $\neg(\varphi \wedge \psi)$ and $\Diamond \varphi \wedge \psi$ and $\varphi \wedge \Diamond \psi$;
\item $PO(X,Y) :$ $\varphi \wedge \psi$ and $\varphi \wedge \neg \psi$ and $\neg \varphi \wedge \psi$; 
\item $TPP(X,Y) :$ $\varphi \Rightarrow \psi$ and $\Diamond \varphi \wedge \neg \psi$;
\item $NTPP(X,Y) :$ $\varphi \Rightarrow \psi$ and $\varphi \Rightarrow \Box \psi$;
\item $EQ(X,Y) : \varphi \Leftrightarrow \psi$.
\end{itemize}
where $\varphi$ and $\psi$ are formulas defining regions $X$ and $Y$, respectively. Let us detail the meaning of some of these formulas:
\begin{itemize}
\item For $EC$: the two regions $X$ and $Y$ do not intersect, but as soon as one of them is dilated (using  $\Diamond$) then they do intersect.
\item For $TPP$: $X$ is included in $Y$ but the dilation of $X$ (represented by $\Diamond \varphi$) is not.
\item For $NTPP$: $X$ is included in $Y$ and also in the erosion of $Y$. 
\end{itemize}

The meaning of the other relations is very natural.

\section{Conclusion}

In this article, we deepened in the topos framework the strong link between MM and modal logic initiated twenty years ago in~\cite{Blo02}. The interest of toposes is that they generalize the notion of space and subspace, and then they include a large family of algebraic structures which have proved useful for knowledge representation and  reasoning.

We have then generalized MM from a topos perspective that we extended by moving from the power object $PX$ to $PPX$, for $X$ an object of a topos $\mathcal{C}$, to define structuring neighborhoods. We then showed that the morphological operations of erosion and dilation have all the properties to make $PX$ an internal modal algebra, and then can be used to give a neighborhood semantics to constructive modal logic in the topos setting. 

To our knowledge, MM has never been studied independently of a given structure. In the complete lattice setting, basic MM operators are defined without referring to the notion of structuring element in the algebraic definitions. Our work is then the first proposition of a generalization of MM independently of a particular structure (set, graph...). Taking into account the notion of structuring element and structuring neighborhood allowed us to generalize the notion of Kripke model and neighborhood model in the topos framework. 

{The proposed extension of basic operators relies on structuring elements defined as morphisms $X \rightarrow PX$. Dilations and erosions using a structuring element are then defined as morphisms $PX \rightarrow PX$ with specific properties. Dilations, erosions, as well as their compositions are shown to have similar properties as in classical morphology (adjunction, monotony...). 
Interestingly enough, the proofs of the main results in this paper highly benefit from the internal logic of topos. {The underlying logic being intuitionistic, as future work, it might be interesting to certify these proofs using a formal proof management system, such as Coq\footnote{https://coq.inria.fr/}.}

Finally, we introduced structuring neighborhoods and topological neighborhoods, giving rise to modal algebras, for which a sound and complete entailment system is proposed.

\Isa{This led us to suggest applications to typical reasoning problems in AI, such as revision, merging, abduction, spatial reasoning based on RCC-8 relations.}

{Let us now mention some other directions for future work.}

\paragraph{Towards a unification of morphological operators via the notion of predicate lifting.} By the bijection correspondence $\Hom(X,PY) \simeq \Sub(X \times Y)$, \Isa{structuring elements and structuring neighborhoods} can be seen as mappings which match each element $x$ with elements in relation with it. A generalization of that can be obtained via the notion of co-algebras~\cite{Rut00}.  In computing science, co-algebras have been studied in $\Set$ \Isa{(the category of sets)}, and in this framework the notion of predicate liftings~\cite{Pat03} has been identified as the concept underlying modal operator semantics.  This notion of predicate lifting can be extended to arbitrary topos. Indeed,  suppose a topos $\mathcal{C}$ and a functor $T : \mathcal{C} \to \mathcal{C}$. A {\em predicate lifting} is then a natural transformation $\lambda : \mathcal{P} \Rightarrow \mathcal{P} \circ T$ (recall that $\mathcal{P}$ is the contravariant functor presented in Section~\ref{preliminaries}).
A structuring element $b$ is then defined as a morphism (co-algebra) $b : X \to T(X)$. If we define the morphism $b^{-1}: PT(X) \to PX$ as follows:
$$b^{-1}(Y) = \{x:X \mid \exists y. y \in_{T(X)} Y \wedge b(x) = y\}$$
a {\em morphological operator} $op[b] : PX \to PX$ is then defined by:  $op[b] = b^{-1} \circ \lambda_X$.  
From there, a morphological operator $op[b]$ is an {\em erosion} (respectively a {\em dilation}) if $\lambda_X$ commutes with $\wedge$ (respectively $\vee$). 

Studying this unifying framework in more depth should allow us to generalize the theory of co-algebras,  to other toposes than the category $\Set$, but also to go further in the generalization of MM and its application to modal logic. Among other things, this will allow us to study concepts such as bisimulation and associated preservation results which have been extensively studied in the theory of co-algebras, at a more abstract level.

\paragraph{Extension to the fuzzy case.} It is usual to introduce uncertainty in qualitative spatial reasoning. Then, as future work, we propose to extend our approach to the fuzzy case. This will f irstrequire to introduce fuzziness in a topos. There are some works showing that fuzzy sets do not form toposes in general~\cite{Bar86,Eyt81,Pit82}, but there are very few which extend toposes to the fuzzy case. Actually, up to our knowledge, this has only been proposed in~\cite{Pue98}. We can then rely on this work to introduce fuzziness in toposes. This will allow us to extend erosion and dilation based on structuring elements to the fuzzy case {in this framework}. In the set theory framework, several definitions of MM on fuzzy sets with structuring elements have been proposed in the literature, since the early work in~\cite{Bae95,BM93} (see e.g~\cite{IB:FSS-09,BM95,NACH-00} for reviews). We will be able to take inspiration from the approach developed in~\cite{IB:FSS-09,IB:IJAR-12,sussner2016} using conjunctions and implications from a topos perspective. {More generally, we could replace the traditional $[0,1]$ interval by any lattice $L$, thus opening the way towards bipolar information (positive as preferences, arguments, observations, and negative as constraints, attacks, etc.). This could find applications in preference-based reasoning, argumentation, spatial reasoning, among others.}

\paragraph{Pretopos structures.} 
{There are spatial structures which are not toposes but only pretoposes. The most emblematic example is the category of compact Hausdorff spaces. Now the properties of pretoposes (exactness and extensivity) make them a very set-like category. Among others, they have an internal logic. The difficulty for extending our work to such structures will be to see how to get around the absence of the power object, a notion on which many of the concepts presented in the paper are built.}



\bibliographystyle{plain}
\bibliography{biblio}

\appendix

\section{Logic and Internal Language in Topos}

An interesting feature of toposes
is that we can reason on objects and morphisms of a topos ``as if they were sets and functions"~\cite{Johnstone02,Law72}. The reason is that we can do logic in toposes. Indeed, we can define logical connectives in toposes. Here, we recall the definition of propositional connectives 
$\{\wedge,\vee,\neg,\Rightarrow\}$ and of constants $true, false$. 
\begin{itemize}
    \item By definition of subobject classifiers, we have a monomorphism $true : \mathbb{1} \rightarrowtail \Omega$, and then we also have a morphism $(true,true) : \mathbb{1} \rightarrowtail \Omega \times \Omega$ which is also a monomorphism. So, by the subobject classifier definition, $\wedge : \Omega \times \Omega \to \Omega$ is its characteristic morphism. 
    \item {$\vee : \Omega \times \Omega \to \Omega$ classifies the image of the morphism $[(true,Id_\Omega),(Id_\Omega,true)] : \Omega +\Omega \to \Omega \times \Omega$,} {where $+$ denotes the co-product.}
    \item the morphism $\Rightarrow : \Omega \times \Omega \to \Omega$ is the characteristic morphism of $\preceq \rightarrowtail \Omega \times \Omega$ where $\preceq$ is the equalizer of $\wedge$ and the projection on the first argument $p_1 : \Omega \times \Omega \to \Omega$. 
    \item Finally, the unique morphism $\emptyset \rightarrowtail \mathbb{1}$ is a monomorphism. Let us denote by $false : \mathbb{1} \to \Omega$ its characteristic morphism. Then, $\neg : \Omega \to \Omega$ is defined as the composite $\Rightarrow \circ \, (Id_\Omega \times false)$. 
\end{itemize}
Consequently the power object $\Omega = P\, \mathbb{1}$ is an internal Heyting algebra\footnote{An {\em internal Heyting algebra} in a topos is an internal lattice $L$, that is equipped with two morphisms $\wedge,\vee : L \times L \to L$ such that the diagrams expressing the standard laws for $\wedge$ and $\vee$ commute, and with top and bottom which are morphisms $\bot,\top: \mathbb{1} \to L$ such that $\wedge\circ (Id_L \times \top) = Id_L$ and $\vee \circ (Id_L \times \bot) = Id_L$, together with an additional morphism $\Rightarrow : L \times L \to L$ which satisfies the diagrams given by the
identities:
\begin{itemize}
    \item $x \Rightarrow x = \top$
    \item $x \wedge (x \Rightarrow y) = x \wedge y$ and $y \wedge (x \Rightarrow y) = y$
    \item $x \Rightarrow (y \wedge z) = (x \Rightarrow y) \wedge (x \Rightarrow z)$
\end{itemize}
}
and then the logic is intuitionistic. Actually, through the bijection $\Sub(X \times Y) \simeq \Hom_\mathcal{C}(X,PY)$, for every object $X$ in a topos $\mathcal{C}$, $PX$ is an internal Heyting algebra. We can then define a partial order $\preceq_X$ 
as an object of $\mathcal{C}$ such that $\preceq_X$ is the equalizer of $\wedge : PX \times PX \to PX$ and $p_1 : PX \times PX \to PX$ where $p_1$ is the projection on the first argument of couples. 

For every topos $\mathcal{C}$, we can define an internal language $\mathcal{L}_\mathcal{C}$ composed of types defined by the objects of $\mathcal{C}$, from which we can define terms as follows:

\begin{itemize}
\item $true:X$;
\item $x:X$ where $x$ is a variable and $X$ is a type;
\item $f(t):Y$ where $f : X \to Y$ is a morphism of $\mathcal{C}$ and $t:X$ is a term;
\item $<t_1,\ldots,t_n>: X_1 \times \ldots \times X_n$ if for every $i$, $1 \leq i \leq n$, $t_i:X_i$ is a term;
\item $(t)_i:X_i$ if $t:X_1 \times \ldots \times X_n$ is a term;
\item $\{x:X \mid \alpha\}:PX$ if $\alpha:\Omega$ is a term;
\item $\sigma = \tau:\Omega$ if $\sigma$ and $\tau$ are terms of the same type;
\item $\sigma \in_X \tau:\Omega$ if $\sigma:X$ and $\tau:PX$ are terms;
\item $\sigma \preceq_X \tau:\Omega$ if $\sigma,\tau:PX$ are terms;
\item $\varphi~@~\psi:\Omega$ if $\varphi:\Omega$ and $\psi:\Omega$ are terms with $@ \in \{\wedge,\vee,\Rightarrow\}$;
\item $\neg \varphi:\Omega$ if $\varphi:\Omega$ is a term;
\item $Q x.\,\varphi:\Omega$ if $x:X$ and $\varphi:\Omega$ are terms and $Q \in \{\forall,\exists\}$.
\end{itemize}
Terms of type $\Omega$ are called {\em formulas}.

Semantics of terms will depend on their type. Hence, semantics of terms of type $X \neq \Omega$ will be defined by morphisms, and terms of type $\Omega$ will be interpreted as subobjects. 

We say that a sequence of variables $\vec{x} = (x_1,\ldots,x_n)$ is a {\em suitable context} for a term or a formula if each free variable of this term or this formula occurs in $\vec{x}$. Let us denote by $X_{\vec{x}}$ the product $X_1 \times \ldots \times X_n$ when $\vec{x} = (x_1,\ldots,x_n)$ and each $x_i : X_i$. Then the {\em semantics} of $t:X$ in the context $\vec{x}$, denoted by $\sem{t}_{\vec{x}}$, is a morphism from $X_{\vec{x}}$ to $X$.  It is defined recursively on the structure of $t$ as follows:

\begin{itemize}
\item $\sem{x_i:X_i}_{\vec{x}} = p_i$ where $p_i : X_{\vec{x}} \to X_i$ is the obvious projection on the $i^{th}$ argument;
\item $\sem{f(t)}_{\vec{x}} = f \circ \sem{t}_{\vec{x}}$;
\item $\sem{<t_1,\ldots,t_n>}_{\vec{x}} = (\sem{t_1}_{\vec{x}},\ldots,\sem{t_n}_{\vec{x}})$;
\item $\sem{(t)_i}_{\vec{x}} = p_i \circ \sem{t}_{\vec{x}}$ where $p_i$ is the projection on the $i^{th}$ argument of the tuple;
\item $\sem{\{x:X \mid \alpha\}}_{\vec{x}}$ is the unique morphism $r : X_{\vec{x}} \to PX$ making the diagram below a pullback square
$$\xymatrixcolsep{5pc}
\xymatrix{
 R  \ar[r] \ar[d]_{\sem{\alpha}_{(x,\vec{x})}} & \in_X \ar[d] \\
 X \times X_{\vec{x}} \ar[r]^-{Id_X \times r}  & X \times PX
 }
$$
\end{itemize}

The semantics of a formula $\varphi:\Omega$ in the context $\vec{x}$, denoted by $\sem{\varphi}_{\vec{x}}$, is interpreted as a subobject of $\Sub(X_{\vec{x}})$ and is recursively defined as follows:
 
\begin{itemize}
\item $\sem{true}_{\vec{x}} = Id_{X_{\vec{x}}}$; 
\item when $\varphi = \sigma = \tau$, then $\sem{\varphi}_{\vec{x}}$ equalizes $\sem{\sigma}_{\vec{x}}$ and $\sem{\tau}_{\vec{x}}$;
\item when $\varphi = \sigma \in_X \tau$, then $\sem{\varphi}_{\vec{x}} : R \rightarrowtail X_{\vec{x}}$ where $R$ is the pullback of the diagram
$$\xymatrixcolsep{5pc}
\xymatrix{
R  \ar[r] \ar[d]_{\sem{\varphi}_{\vec{x}}} & \in_X \ar[d] \\
 X_{\vec{x}} \ar[r]_-{\sem{\sigma}_{\vec{x}} \times \sem{\tau}_{\vec{x}}} & X \times PX
 }
$$
\item if $\varphi = \sigma \preceq_X \tau$, then $\sem{\varphi}_{\vec{x}} : R \rightarrowtail X_{\vec{x}}$  where $R$ is the pullback of the diagram 
$$\xymatrixcolsep{5pc}
\xymatrix{
R  \ar[r] \ar[d]_{\sem{\varphi}_{\vec{x}}} & \preceq_X \ar[d] \\
 X_{\vec{x}} \ar[r]_-{\sem{\sigma}_{\vec{x}} \times \sem{\tau}_{\vec{x}}} & PX \times PX
 }
$$
\item if $\varphi = \varphi_1~@~\varphi_2$, then $\sem{\varphi}_{\vec{x}} = \sem{\varphi_1}_{\vec{x}}~@~\sem{\varphi_2}_{x_t}$ where $@$ is the operator in $\{\wedge,\vee,\Rightarrow\}$ in the Heyting algebra $\Sub(X_{\vec{x}})$;
\item $\sem{\neg \varphi}_{\vec{x}} = \neg_{X_{\vec{x}}} (\sem{\varphi}_{\vec{x}})$ where $\neg_{X_{\vec{x}}}(\sem{\varphi}_{\vec{x}})$ is the pseudo-complement of $\sem{\varphi}_{\vec{x}}$ in $\Sub(X_{\vec{x}})$;
\item $\sem{\forall x.\varphi}_{\vec{x}} = \forall_p(\sem{\varphi}_{(\vec{x},x)})$ where $p : X_{\vec{x}} \times X \to X_{\vec{x}}$ is the projection, and $\forall_p$ is the right adjoint to the pullback functor $p^* : \Sub(X_{\vec{x}}) \to \Sub(X_{\vec{x}} \times X)$ when the Heyting algebras $\Sub(X_{\vec{x}})$ and $\Sub(X_{\vec{x}} \times X)$ are regarded as categories. 
\item $\sem{\exists x.\varphi}_{\vec{x}}$ is the image of $p \circ \sem{\varphi}_{(\vec{x},x)}$ where $p$ is the same projection as above.
\end{itemize}
Equivalently, semantics of any formula $\varphi:\Omega$ could be defined by a morphism from $X_{\vec{x}}$ to $\Omega$, by interpreting $\varphi$ as the classifying morphism of $\sem{\varphi}_{\vec{x}}$. 

We write $\mathcal{C} \models_{\vec{x}} \varphi$ if $\sem{\varphi}_{\vec{x}} = Id_{X_{\vec{x}}}$ ($Id_{X_{\vec{x}}}$ is the top element in $\Sub(X_{\vec{x}})$). However, to actually prove properties on toposes, the best way is to use the internal logic. This internal logic is defined as for any logic by a set of inference rules which allow us to derive true statements from other true statements. In order to describe this better we need to introduce the notion of a sequent.

\begin{definition}[Sequent]
Given two formulas $\varphi$ and $\psi$, a {\bf sequent} is an expression of the form $\varphi \vdash_{\vec{x}} \psi$ and means that $\psi$ is a logical consequence of $\varphi$ in the context $\vec{x}$.
\end{definition}
{Note that if the sequence $\vec{x}$ is empty, then we will note simply $\vdash$ and $\models$.}

The deduction system is then defined as a sequent calculus, i.e. a set of inference rules which will allow us to infer a sequent from other sequents (see \cite{Johnstone02,MM12,Osi75} for a presentation of this deduction system). 

\section{Proofs of Section~\ref{results standard MM}}

\subsection{Proof of Proposition~\ref{adjunction} (adjunction)}

{Let us first recall a useful lemma.}
\begin{lemma}
\[
\varphi \Rightarrow \psi 
\ssequent
\varphi \wedge \psi \Leftrightarrow \varphi
\]
\end{lemma}
{where $A \ssequent B$ is a shortened notation for $A \vdash B$ and $B \vdash A$.}

The result in Proposition~\ref{adjunction} then rests on the two following propositions.

\begin{proposition}
\label{order and membership}
$$\vdash_{\vec{x}} Y \preceq_X Z \Leftrightarrow  (\forall x.\; x \in_X Y \Rightarrow x \in_X Z)$$
\end{proposition}



\begin{proof}
\begin{align*}
Y \preceq_X Z
&\ssequent_{Y,Z}
Y \wedge Z = Y
\\
&\ssequent_{Y,Z,x}
x \in_X\! Y \wedge x \in_X\! Z \Leftrightarrow x \in_X\! Y
\\
&\ssequent_{Y,Z,x} x \in_X\! Y \Rightarrow x\in_X\! Z
\\
&\ssequent_{Y,Z} \forall x.\; x\in_X\! Y \Rightarrow x\in_X\! Z
\end{align*}
\end{proof}

\begin{proposition}
\label{dilation union of b}
The following statement is valid:
$$\vdash \forall y.\; y \in_X\! Y \Rightarrow (\forall x.\; x \in_X\! b(y) \Rightarrow x \in_X\! \delta[b](Y)) $$ 
\end{proposition}

\begin{proof}
By using the internal logic, by both universal quantification and implication rules, this amounts to prove that
$$y \in_X\! Y \wedge y \in_X\! \breve{b}(x) \vdash_{Y,y,x} x \in_X\! \delta[b](Y)$$
(We use the fact that $x \in_x b(y) \Leftrightarrow y \in_X\! \breve{b}(x)$).
{The existential quantification rule combined with the definition of dilation concludes.} 
{Indeed,} by the existential quantification rule, we have
$$\exists y.\; y \in_X\! Y \wedge y \in_X\! \breve{b}(x) \vdash_{Y,x} true$$
and then, by introduction of hypothesis, we can write
$$\exists y.\; y \in_X\! Y \wedge y \in_X\! \breve{b}(x) \vdash_{Y,x} \exists y.\; y \in_X\! Y \wedge y \in_X\! \breve{b}(x)$$
Then, by the definition of dilation, we have
$$\exists y.\; y \in_X\! Y \wedge y \in_X\! \breve{b}(x) \vdash_{Y,x} x \in_X\! \delta[b](Y)$$
By using the existential quantification rule, this is equivalent to
$$y \in_X\! Y \wedge y \in_X\! \breve{b}(x) \vdash_{Y,y,x} x \in_X\! \delta[b](Y)$$ 
\end{proof}

{Note that he opposite implication $\Leftarrow$ in Proposition~\ref{dilation union of b} is also valid.}

{It is now easy to prove the adjunction property. Its proof is substantially similar to that above.}

\begin{proof} {of Proposition~\ref{adjunction}:}
By definition, both dilation and erosion of any object $Y:PX$ define subobjects $\varepsilon[b](Y) \rightarrowtail X$ and $\delta[b](Y) \rightarrowtail X$. {Dilation and erosion both result in applications $\Sub(X) \to \Sub(X)$.} Moreover, by proposition~\ref{dilation union of b} we have that
$$\vdash_{Y} \forall y.\; y \in_X\! Y \Rightarrow (\forall x.\; x \in_X\! b(y) \Rightarrow x \in_X\! \delta[b](Y))$$
From Proposition~\ref{order and membership}, we deduce that 
$$\vdash_{Y}  \forall y.\; y \in_X\! Y \Rightarrow b(y) \preceq_X \delta[b](Y)$$ 
Then, we have the following equivalences:
$$
\begin{array}{llr}
\vdash_{Y,Z} \delta[b](Y) \preceq_X Z & \Leftrightarrow \forall y.\; y \in_X\! Y \Rightarrow b(y) \preceq_X Z & \\
                                     			  & \Leftrightarrow \forall y.\; y \in_X\! Y \Rightarrow y \in_X \varepsilon[b](Z) & \mbox{(definition of erosion)}\\
                                      			  & \Leftrightarrow Y \preceq_X \varepsilon[b](Z) & \mbox{(Proposition~\ref{order and membership})}
\end{array}
$$ 
\end{proof}

\subsection{Proof of Proposition~\ref{extensivity, etc.}}

Let us show the properties for erosion. Let us show first {the monotony property}:
$$\vdash_{Y,Z}  Y \preceq_X Z \Longrightarrow \varepsilon[b](Y) \preceq_X \varepsilon[b](Z)$$
We have the following implications:
$$ 
\begin{array}{llr}
Y \preceq_X Z \vdash_{Y,Z} \forall x.\; x \in_X\! \varepsilon[b](Y) & \Rightarrow b(x) \preceq Y & \mbox{ (definition of erosion)} \\
		                  							        & \Rightarrow b(x) \preceq Z & \mbox{ (hypothesis and transitivity of $\preceq_X$)} \\
				                 					        & \Rightarrow x \in_X\! \varepsilon[b](Z) & \mbox{ (definition of erosion)} \\
				          					                & \Rightarrow \varepsilon[b](Y) \preceq_X \varepsilon[b](Z) & \mbox{ (Proposition~\ref{order and membership})}
\end{array}
$$

Let us show the {commutativity of erosion with conjunction}: 
$$\vdash \forall Y.\; \forall Z.\; \varepsilon[b](Y \wedge Z) = \varepsilon[b](Y) \wedge \varepsilon[b](Z)$$
By definition of erosion, we have that
$$\vdash_{Y,Z} \forall x.\; x \in_X\! \varepsilon[b](Y \wedge Z) \Leftrightarrow b(x) \preceq_X Y \wedge Z$$
and then we have that 
$$\vdash_{Y,Z} \forall x.\; x \in_X\! \varepsilon[b](Y \wedge Z) \Leftrightarrow b(x) \preceq_X Y \wedge b(x) \preceq_X Z$$ 
By definition of erosion, we can then derive that 
$$\vdash_{Y,Z} \forall x.\; x \in_X\! \varepsilon[b](Y \wedge Z) \Leftrightarrow x \in_X\! \varepsilon[b](Y) \wedge x \in_X\! \varepsilon[b](Z)$$
that is:
$$\vdash_{Y,Z} \forall x.\; x \in_X\! \varepsilon[b](Y \wedge Z) \Leftrightarrow x \in_X\! \varepsilon[b](Y) \wedge \varepsilon[b](Z)$$
\Isa{This extends to any conjunction.}

Now, let us show that $\vdash \varepsilon[b](X) = X$. By definition of erosion, we have that
$$\vdash \forall x.\; x \in_X \varepsilon[b](X) \Leftrightarrow b(x) \preceq_X X$$
Now, by the topos properties, we have that 
$$\vdash \forall x.\; b(x) \preceq_X X$$
and then, by the cut rule, we can write 
$$\vdash \forall x.\; x \in_X\! \varepsilon[b](X)$$
from which we derive
$$\vdash \varepsilon[b](X) = X$$

To finish, {let us prove the necessary and sufficient condition for anti-extensivity of erosion, and first} show that:
$$\vdash (\forall x.\; x \in_X\! b(x)) \Rightarrow (\forall Y.\; \varepsilon[b](Y) \preceq_X\! Y)$$
By the rules for the implication connective and universal quantifier, this is equivalent to prove the following statement:
$$\forall x.\; x \in_X b(x) \vdash_{Y} \varepsilon[b](Y) \preceq_X Y$$
By introduction of hypothesis, we have
$$\forall x.\; x \in_X b(x) \vdash_{Y} \forall x.\; x \in_X\! b(x)$$
and then, by the rule for the universal quantifier, this leads to
$$\forall x.\; x \in_X\! b(x) \vdash_{Y,x} x \in_X\! b(x)$$
By the thinning rule, we have
$$\forall x.\; x \in_X b(x), \; x \in_X\! \varepsilon[b](Y) \vdash_{Y,x} x \in_X\! b(x)$$
Now, by the introduction of hypothesis and the definition of $\varepsilon[b]$, we have
$$\forall x.\; x \in_X\! b(x), \; x \in_X\! \varepsilon[b](Y) \vdash_{Y,x} b(x) \preceq_X Y$$
and then by Proposition~\ref{order and membership}, we deduce
$$\forall x.\; x \in_X\! b(x),\; x \in_X\! \varepsilon[b](Y) \vdash_{Y,x} x \in_X\! Y$$
By the rule for the implication, we then have
$$\forall x.\; x \in_X\! b(x) \vdash_{Y,x} x \in_X\! \varepsilon[b](Y) \Rightarrow x \in_X\! Y$$
which gives, by the rule for the universal quantifier:
$$\forall x.\; x \in_X\! b(x) \vdash_{Y} \forall x.\;  x \in_X\! \varepsilon[b](Y) \Rightarrow x \in_X\! Y$$
which, by Proposition~\ref{order and membership}, allows us to conclude
$$\forall x.\; x \in_X\! b(x) \vdash_{Y} \varepsilon[b](Y) \preceq_X Y$$

Let us now prove the opposite sequent: we have
\[
    \forall Y.\; \varepsilon[b](Y) \preceq_X Y 
     \vdash_{Y} 
    \varepsilon[b](Y) \preceq_X Y
\]
and then by substitution:
\[
    \forall Y.\; \varepsilon[b](Y) \preceq_X Y 
     \vdash_{x} 
    \varepsilon[b](b(x)) \preceq_X b(x)
\]
We have by definition $\vdash_{x} x \in_X \varepsilon[b](b(x))$, so by thinning and finite conjunction rules, we have:
\begin{align*}
    \forall Y.\; \varepsilon[b](Y) \preceq_X Y 
     &\vdash_{x} 
     x \in_X\! \varepsilon[b](b(x))
     \wedge 
     \varepsilon[b](b(x)) \preceq_X b(x)
     \\
     &\vdash_{x} x \in_X\! b(x)
\end{align*}
which concludes by the universal quantification rule:
$$\forall Y.\; \varepsilon[b](Y) \preceq_X Y \vdash_{Y} \forall x. \;x \in_X\! b(x)$$

The proofs for dilation are substantially similar.

 {Let us only prove the necessary and sufficient condition for extensivity of dilation.}
{We have:}
\begin{align*}
(\forall x. \;x \in_X\! b(x)) \; \wedge \; x \in_X\! Y 
&\vdash_{Y,x} x \in_X\! b(x) \wedge x \in_X\! Y
\\
&\vdash_{Y,x,y} y=x \wedge y \in_X\! \breve b(x) \wedge y \in_X\! Y
\\
&\vdash_{Y,x} \exists y.\; y=x \wedge y \in_X\! \breve b(x) \wedge y \in_X\! Y
\\
&\vdash_{Y,x} \exists y.\; y \in_X\! \breve b(x) \wedge y \in_X\! Y
\\
&\vdash_{Y,x} x \in \delta[b](Y)
\end{align*}
and by implication and universal quantification rules, we derive:
\[
\forall x. \;x \in_X\! b(x)
\vdash
\forall Y. \; Y \preceq_X \delta[b](Y)
\]
Conversely:
\begin{align*}
\forall Y. \; Y \preceq_X \delta[b](Y)
&\vdash_x
\{x\} \preceq_X \delta[b](\{x\})
\\
&\vdash_x
x \in_X\! \delta[b](\{x\})
\\
&\vdash_x
\exists y.\; x \in_X\! b(y) \wedge y \in_X\! \{x\}
\\
&\vdash_x
\exists y.\; x \in_X\! b(x)
\\
&\vdash_x
x \in_X\! b(x)
\end{align*}
and by universal quantification:
\[
\forall Y. \; Y \preceq_X \delta[b](Y)
\vdash
\forall x. \; x \in_X\! b(x)
\]

\subsection{Proof of Proposition~\ref{opening and closing}}

\begin{itemize}
\item \Isa{{\em Monotony} is derived directly from the monotony of $\varepsilon[b]$ and $\delta[b]$.}
\item {\em Extensivity of closing.} Obviously, we have that 
$$\vdash_{Y} \delta[b](Y) \preceq_X \delta[b](Y)$$ 
and then by the adjunction property, we can conclude that 
$$\vdash_{Y} Y \preceq_X \varepsilon[b](\delta[b](Y))$$
\item {\em Anti-extensivity of opening.} Obviously, we have that 
$$\vdash_{Y} \varepsilon[b](Y) \preceq_X \varepsilon[b](Y)$$ 
and then by the adjunction property, we can conclude that 
$$\vdash_{Y} \delta[b](\varepsilon[b](Y)) \preceq_X Y$$
\item {\em Preservation.} By anti-extensivity of opening we have that 
$$\vdash_{Y} \delta[b](\varepsilon[b](Y)) \preceq_X Y$$ 
and then by monotonicity of erosion, we can conclude that
$$\vdash_{Y} \varepsilon[b](\delta[b](\varepsilon[b](Y))) \preceq_X \varepsilon[b](Y)$$
To show the opposite direction, we obviously have that 
$$\vdash_{Y} \delta[b](\varepsilon[b](Y)) \preceq_X \delta[b](\varepsilon[b](Y))$$ 
and then by the adjunction property, we can conclude that 
$$\vdash_{Y} \varepsilon[b](Y) \preceq_X \varepsilon[b](\delta[b](\varepsilon[b](Y)))$$ 
The equation $\delta[b] \circ \varepsilon[b] \circ \delta[b] = \delta[b]$ can be proved similarly.
\item {\em Idempotence.} Direct consequence of the preservation properties.
\end{itemize}

\subsection{Proof of Proposition~\ref{prop:minimality}}

$(\Longrightarrow)$ Let us suppose that the revision operator $\circ$ satisfies AGM Postulates. Let us define the binary relation $\preceq^\varphi_\psi$ on $Mod(\psi)$ as follows:
$$\mathcal{M} \preceq^\varphi_\psi \mathcal{M}' \; \mbox{iff} \; \mathcal{M} \in Mod(\varphi \circ \psi) \; \mbox{and} \; \mathcal{M}' \notin Mod(\varphi \circ \psi)$$
Let us set $f(t) = \preceq_\varphi = \bigcup_{\psi} \preceq^\varphi_\psi$. Let us first show that $\preceq_\varphi$ is a FA. Then, let us show that the mapping $f$ is a FA. 
\begin{itemize}
    \item Let $\mathcal{M},\mathcal{M}' \in Mod(\varphi)$, and let us suppose that $\mathcal{M} \preceq_\varphi \mathcal{M}'$. This means that there exists $\psi$ such that $\mathcal{M} \preceq^\varphi_\psi \mathcal{M}'$, i.e. $\mathcal{M},\mathcal{M}' \in Mod(\psi)$, $\mathcal{M} \in Mod(\varphi \circ \psi)$ and $\mathcal{M}' \notin Mod(\varphi \circ \psi)$. Hence, we have that $\varphi \wedge \psi$ is consistent, and then by Postulate $(G3)$, $\varphi \circ \psi = \varphi \wedge \psi$. We then have that $\mathcal{M}' \in Mod(\varphi \circ \psi)$ which is a contradiction.
    \item Let $\mathcal{M} \in Mod(\varphi)$ and let $\mathcal{M}' \notin Mod(\varphi)$. We have that $\mathcal{M} \preceq^\varphi_\top \mathcal{M}'$ by definition of $\preceq_\varphi$. Now, let us suppose that $\mathcal{M}' \preceq_\varphi \mathcal{M}$. This means that there exists $\psi$ such that $\mathcal{M}' \preceq^\varphi_\psi \mathcal{M}$. Hence, we have that $\mathcal{M},\mathcal{M}' \in Mod(\psi)$, $\mathcal{M}' \in Mod(\varphi \circ \psi)$ and $\mathcal{M} \notin Mod(\varphi \circ \psi)$, and then as $\mathcal{M} \in Mod(\varphi)$ we have that $\varphi \wedge \psi$ is consistent. By Postulate $(G3)$, we then have that $\varphi \circ \psi = \varphi \wedge \psi$, and then $\mathcal{M} \in Mod(\varphi \circ \psi)$ which is a contradiction.
\end{itemize}

Let us show now that $Mod(\varphi \circ \psi) = Min(Mod(\psi),\preceq_\varphi)$.
\begin{itemize}
    \item Let $\mathcal{M} \in Mod(\varphi \circ \psi)$, and let us suppose that $\mathcal{M} \notin Min(Mod(\psi),\preceq_\varphi)$. By Postulate $(G2)$, $\mathcal{M} \in Mod(\psi)$. By hypothesis, there exists $\mathcal{M}' \in Mod(\psi)$ such that $\mathcal{M}' \prec_\varphi \mathcal{M}$. This means that there exists a formula $\delta$ such that $\mathcal{M}' \in Mod(\varphi \circ \delta)$ and $\mathcal{M} \notin Mod(\varphi \circ \delta)$. By definition of the relation $\preceq^\varphi_\delta$, we have that $\psi \wedge \delta$ is consistent, and then by Postulate $(G5)$, we can write
    $$Mod((\varphi \circ \psi) \wedge \delta) = Mod((\varphi \circ \delta) \wedge \psi) = Mod(\varphi \circ (\psi \wedge \delta))$$
    Hence, we have that $\mathcal{M} \in Mod(\varphi \circ \delta)$ which is a contradiction.
    \item Let $\mathcal{M} \in Min(Mod(\psi),\preceq_\varphi)$. Let us assume that $\mathcal{M} \notin Mod(\varphi \circ \psi)$. Let us suppose $\mathcal{M}' \in Mod(\varphi \circ \psi)$. By definition of $\preceq^\varphi_\psi$, we have that $\mathcal{M}' \preceq^\varphi_\psi \mathcal{M}$, and then $\mathcal{M}' \preceq_\varphi \mathcal{M}$. Now, let us suppose that $\mathcal{M} \preceq_\varphi \mathcal{M}'$. This means that there exists $\delta$ such that $\mathcal{M} \in Mod(\varphi \circ \delta)$ and $\mathcal{M}' \notin Mod(\varphi \circ \delta)$. But, $\mathcal{M}' \in Mod(\psi \wedge \delta)$, and by Postulate $(G5)$ we have that 
    $$Mod((\varphi \circ \psi) \wedge \delta) = Mod((\varphi \circ \delta) \wedge \psi) = Mod(\varphi \circ (\psi \wedge \delta))$$
    Therefore, we have that $\mathcal{M} \in Mod(\varphi \circ \delta)$ which is a contradiction. Hence, we just showed that $\mathcal{M}' \prec_\varphi \mathcal{M}$ which is a contradiction.
\end{itemize}

$(\Longleftarrow)$ Let us now suppose that for a revision operation $\circ$ there exists a FA which maps any formula $\varphi$ to a binary relation $\preceq_\varphi$ satisfying the conditions of Proposition~\ref{prop:minimality}. Let us prove that $\circ$ satisfies the AGM Postulates.
\begin{itemize}
    \item $(G1)$ As $\psi$ is consistent, we have that $Min(Mod(\psi),\preceq_\varphi) \neq \emptyset$, and then by the hypothesis $\varphi \circ \psi$ is consistent.
    \item $(G2)$ Let $\mathcal{M} \in Mod(\varphi \circ \psi)$. By hypothesis, we have that 
    $$\mathcal{M} \in Min(Mod(\psi),\preceq_\varphi)$$ 
    and then $\mathcal{M} \in Mod(\psi)$.
    \item $(G3)$ Let us suppose that $\varphi \circ \psi$ is consistent. 
    \begin{itemize}
        \item Let us prove that $Mod(\varphi \circ \psi) \subseteq Mod(\varphi \wedge \psi)$. Let $\mathcal{M} \in Mod(\varphi \circ \psi)$. Hence, we have that $\mathcal{M} \in Mod(\psi)$. Let us suppose that $\mathcal{M} \notin Mod(\varphi)$. Let $\mathcal{M}' \in Mod(\varphi)$. By the definition of FA, we have that $\mathcal{M}' \prec_\varphi \mathcal{M}$ which contradicts the fact that $\mathcal{M}\in Min(Mod(\psi),\preceq_\varphi)$.
        \item Let us prove that $Mod(\varphi \wedge \psi) \subseteq Mod(\varphi \circ \psi)$. Let $\mathcal{M} \in Mod(\varphi \wedge \psi)$ be a model such that $\mathcal{M} \notin Mod(\varphi \circ \psi)$. We have that $\mathcal{M} \in Mod(\psi)$. By hypothesis, there exists a model $\mathcal{M}' \in Mod(\psi)$ such that $\mathcal{M}' \prec_\varphi \mathcal{M}$, and then $\mathcal{M} \notin Mod(\varphi)$ which is a contradiction.  
    \end{itemize}
    \item $(G4)$ Let us suppose that $\psi \equiv \psi'$. This means that $Mod(\psi) = Mod(\psi')$, and then $Min(Mod(\psi),\preceq_\varphi) = Min(Mod(\psi'),\preceq_\varphi)$. We can then conclude that $Mod(\varphi \circ \psi) = Mod(\varphi \circ \psi')$. 
    \item $(G5)$ \Isa{Let us suppose that $(\varphi \circ \psi) \wedge \chi$ is consistent.}
    \begin{itemize}
        \item Let us suppose that $\mathcal{M} \in Mod((\varphi \circ \psi) \wedge \chi)$. Let us suppose that $\mathcal{M} \notin Min(\varphi \circ (\psi \wedge \chi))$. This means that there exists $\mathcal{M}' \in Mod(\psi \wedge \chi)$ such that $\mathcal{M}' \prec_\varphi \mathcal{M}$. We have then that $\mathcal{M}' \in Mod(\psi)$, and then $\mathcal{M} \prec_\varphi \mathcal{M}'$ which is a contradiction. 
        \item Let us suppose that $\mathcal{M} \in Mod(\varphi \circ (\psi \wedge \chi))$. We then have that $\mathcal{M} \in Mod(\psi \wedge \chi)$. Now, we have that 
        $$Min(Mod(\psi \wedge \chi),\preceq_\varphi) = Min(Mod(\psi),\preceq_\varphi) \cap Mod(\chi)$$ 
        and then $\mathcal{M} \in Mod((\varphi \circ \psi) \wedge \chi)$.
    \end{itemize}
\end{itemize}

\end{document}